\documentclass[12, a4paper]{article}
\usepackage[utf8]{inputenc}

\usepackage[english]{babel}

\usepackage[explicit]{titlesec}
\usepackage{blindtext}

\usepackage{lineno}
\usepackage{geometry}
\usepackage{setspace}
\usepackage{indentfirst}
\usepackage{csquotes}

\usepackage[hidelinks]{hyperref}
\usepackage[backend=biber, style=apa, sorting=nyt, uniquelist=false, dashed=false, bibencoding=utf8]{biblatex}
\usepackage{fancyhdr}
\usepackage{graphicx}
\usepackage{float}
\usepackage[center]{caption}
\usepackage{subcaption}
\usepackage{array}
\usepackage{makecell}

\usepackage{tabularx,ragged2e,booktabs,caption,tabu}
\newcolumntype{L}[1]{>{\arraybackslash\raggedright}p{#1}}
\usepackage{longtable}
\usepackage{multirow}
\usepackage[table,xcdraw]{xcolor}
\newcolumntype{P}[1]{>{\centering\arraybackslash}p{#1}}
\newcolumntype{M}[1]{>{\centering\arraybackslash}m{#1}}
\newcolumntype{C}[1]{ >{\centering\arraybackslash} m{3cm} }
\usepackage{placeins}
\usepackage{rotating}
\usepackage{hhline}

\usepackage{authblk} 
\usepackage{pgfplots, pgfplotstable}
\usepackage{tikz}
\usetikzlibrary{shapes.geometric, arrows}
\usetikzlibrary{mindmap,trees,positioning}
\usepackage{smartdiagram}
\usesmartdiagramlibrary{additions}
\usepackage{verbatim}
\tikzstyle{bag} = [align=center]
\pgfplotsset{compat=1.9}
\usetikzlibrary{arrows, decorations.markings}

\usepackage{enumitem}
\begin{document}

\title{Weed Recognition using Deep Learning Techniques on Class-imbalanced Imagery}

\def\correspondingauthor{\footnote{Corresponding author email: F.Sohel@murdoch.edu.au}}
\author[1,2]{A S M Mahmudul Hasan}
\author[1,2]{Ferdous Sohel \correspondingauthor{}}
\author[2,3,4]{Dean Diepeveen}
\author[1,5,6]{Hamid Laga}
\author[2] {Michael G.K. Jones}
\affil[1]{Information Technology, Murdoch University, Murdoch, WA 6150, Australia}
\affil[2]{Centre for Crop and Food Innovation, Food Futures Institute, Murdoch University, Murdoch, WA 6150, Australia}
\affil[3]{Department of Primary Industries and Regional Development, Western Australia, South Perth, WA, 6151, Australia}
\affil[4]{Centre for Sustainable Farming Systems, Murdoch University, Murdoch, WA 6150, Australia}
\affil[5]{Centre of Biosecurity and One Health, Harry Butler Institute, Murdoch University, Murdoch University, Murdoch, WA 6150, Australia}
\affil[6]{Phenomics and Bioinformatics Research Centre, University of South Australia, SA, 5000, Australia}

\date{\vspace{-5ex}}

\maketitle
\cleardoublepage
\pagenumbering{arabic}

\begin{center}
    \LARGE{\bfseries{Weed Recognition using Deep Learning Techniques on Class-imbalanced Imagery}}
\end{center}

\begin{abstract}
    Most weed species can adversely impact agricultural productivity by competing for nutrients required by high-value crops. Manual weeding is not practical for large cropping areas. Many studies have been undertaken to develop automatic weed management systems for agricultural crops. In this process, one of the major tasks is to recognise the weeds from images. However, weed recognition is a challenging task. It is because weed and crop plants can be similar in colour, texture and shape which can be exacerbated further by the imaging conditions, geographic or weather conditions when the images are recorded. Advanced machine learning techniques can be used to recognise weeds from imagery. In this paper, we have investigated five state-of-the-art deep neural networks, namely VGG16, ResNet-50, Inception-V3, Inception-ResNet-v2 and MobileNetV2, and evaluated their performance for weed recognition. We have used several experimental settings and multiple dataset combinations. In particular, we constructed a large weed-crop dataset by combining  several smaller datasets, mitigating class imbalance by data augmentation, and using this dataset in benchmarking the deep neural networks. We investigated the use of transfer learning techniques by preserving the pre-trained weights for extracting the features and fine-tuning them using the images of crop and weed datasets. We found that VGG16 performed better than others on small-scale datasets, while ResNet-50 performed better than other deep networks on the large combined dataset. 
    
\end{abstract}


\section{Introduction} 

Weeds in crops compete for water, nutrients, space and light, and may decrease product quality \parencite{iqbal2019investigation}. Their control, using a range of herbicides, constitutes a significant part of current agricultural practices. In Australia weed control costs in grain production is estimated at \$4.8 billion per annum. These costs include weed control and the cost of lost production \parencite{mcleod_2018}.\par

The most widely used methods for controlling weeds are chemical-based, where herbicides are applied at an early growth stage of the crop \parencite{harker2013recent, lopez2011weed}. Although the weeds spread in small patches in crops, herbicides are usually applied uniformly throughout the agricultural field.  While such an approach works reasonably well against weeds, it also affects the crops. A report from the European Food Safety Authority (EFSA) shows that most of the unprocessed agricultural produces contain harmful substances originating from herbicides \parencite{european20202018}.

Recommended rates of herbicide application are expensive and may also be detrimental to the environment. Thus, new methods that can be used to identify weeds in crops, and then selectively apply herbicides on the weeds, or other methods to control weeds, will reduce production costs to the farmers and benefit the environment. Technologies that enable the rapid discrimination of weeds in crops are now becoming available \parencite{tian2020computer}. \par

Recent advances in Deep Learning (DL) have revolutionised the field of Machine Learning (ML). DL has made a significant impact in the area of computer vision by learning features and tasks directly from audio, images or text data without human intervention or predefined rules \parencite{dargan2019survey}.  For image classification, DL methods outperform humans and traditional ML methods in accuracy and speed \parencite{steinberg_2017}. In addition, the availability of computers with powerful GPUs, coupled with the availability of large amounts of labelled data, enable the efficient training of DL models. \par

As for other computer vision and image analysis problems, digital agriculture and digital farming also benefits from the recent advances in deep learning. Deep learning techniques have been applied for weed and crop management, weed detection, localisation and classification, field conditions and livestock monitoring \parencite{kamilaris2018deep}. \par

ML techniques have been used in commercial solutions to combat weeds. \enquote{Robocrop Spot Sprayer} \parencite{machinery_2018} is a video analysis-based autonomous selective spraying system that can identify potatoes grown in carrots, parsnips, onions or leeks.  \enquote{WeedSeeker sprayer} \parencite{trimble_agriculture} is a near-infrared reflectance sensor-based system that detects the green component in the field. The machine sprays herbicides only on the plants while reducing the amount of herbicide. Similar technology is offered by a herbicide spraying system known as \enquote{WEED-IT} \parencite{weed}. It can target all green plants on the soil. A fundamental problem with these systems is that they are non-selective of crops or weeds. Therefore the ability to discriminate between crops and weeds is important. \par

Further development of autonomous weed control systems can be beneficial both economically and environmentally. Labour costs can be reduced by using a machine to identify and remove weeds. Selective spraying can also minimise the amount of herbicides applied \parencite{lameski2018review}. The success of an autonomous weed control system will depend on four core modules: (i) weed detection and recognition, (ii) mapping, (iii) guidance and (iv) weed control \parencite{olsen2019deepweeds}. This paper focuses on the first module: weed detection and recognition, which is a challenging task \parencite{slaughter2008autonomous}. This is because both weeds and crop plants often exhibit similar colours, textures and shapes. Furthermore, the visual properties of both weeds and crop plants can vary depending on the growth stage, lighting conditions, environments and geographical locations \parencite{jensen2020automated,hasan2021survey}. Also, weeds and crops, exhibit high inter-class similarity as well as high intra-class dissimilarity. The lack of large-scale crop weed datasets is a fundamental problem for DL-based solutions.\par

There are many approaches to recognise weed and crop classes from images \parencite{waldchen2018plant}. 
High accuracy can be obtained for weed classification using Deep Learning (DL) techniques \parencite{kamilaris2018deep} whereas \textcite{chavan2018agroavnet} used Convolutional Neural Network (CNN) models to classify weeds and crop plants. \textcite{teimouri2018weed} used DL for the classification of weed species and the estimation of growth stages, with an average classification accuracy of 70\% and 78\% for growth stage estimation.  \par

As a general rule, the accuracy of the methods used for the classification of weed species decreases in multi-class classification when the number of classes is large \parencite{dyrmann2016plant, peteinatos2020weed}. Class-imbalanced datasets also reduce the performance of DL-based classification techniques because of overfitting \parencite{ali2019mfc}. This problem can be addressed using data-level and algorithm-level methods. Data-level methods include oversampling or undersampling of the data. In contrast, algorithm-level methods work by modifying the existing learning algorithms to concentrate less on the majority group and more on the minority classes. The cost-sensitive learning approach is one such approach \parencite{krawczyk2016learning, khan2017cost}. \par

DL techniques have been used extensively for weed recognition, for example \textcite{hasan2021survey} have provided a comprehensive review of these techniques. \textcite{dos2017weed} compared the performance of CNN with Support Vector Machines (SVM), Adaboost – C4.5, and Random Forest models for discriminating soybean plants, soil, grass, and broadleaf weeds. This study shows that CNN can be used to classify images more accurately than other machine learning approaches. \textcite{nkemelu2018deep} report that CNN models perform better than SVM and K-Nearest Neighbour (KNN) algorithms. \par

Transfer learning is an approach that uses the learned features on one problem or data domain for another related problem. Transfer learning mimics classification used by humans, where a person can identify a new thing using previous experience. In deep learning, pre-trained convolutional layers can be used as a feature extractor for a new dataset \parencite{shao2014transfer}. However, most of the well-known CNN models are trained on ImageNet datasets, which contains 1000 classes of objects. That is why, depending on the number of classes in the desired dataset, only the classification layer (fully connected layer) of the models need to be trained again in the transfer learning approach. \textcite{suh2018transfer} applied six CNN models (AlexNet, VGG-19, GoogLeNet, ResNet-50, ResNet-101 and Inception-v3) pre-trained on the ImageNet dataset to classify sugar beet and volunteer potatoes. They reported that these models can achieve a classification accuracy of about 95\% without retraining the pre-trained weights of the convolutional layers. They also observed that the models' performance improved significantly by fine-tuning the pre-trained weights. In the fine-tuning approach, the convolutional layers of the DL models are initialised with the pre-trained weights, and subsequently during the training phase of the model, those weights are retrained for the desired dataset. Instead of training a model from scratch, initialising it with pre-trained weights and fine-tuning them helps the model to achieve better classification accuracy for a new target dataset, and this also saves training time \parencite{girshick2014rich, hentschel2016fine, gando2016fine}. \textcite{olsen2019deepweeds} fine-tuned the pre-trained ResNet-50 and Inception-V3 models to classify nine weed species in their study and achieved an average accuracy of 95.7\% and 95.1\% respectively. In another study, VGG16, ResNet-50 and Inception-V3 pre-trained models were fine-tuned to classify the weed species found in the corn and soybean production system \parencite{ahmad2021performance}. The VGG16 model achieved the highest classification accuracy of 98.90\% in their research. \par

In this paper, we have performed several experiments: i) First, we evaluated the performance of DL models under the same experimental conditions using small-scale public datasets. ii) We then constructed a large dataset by combining a few small-scale datasets with a variety of weeds in crops. In the dataset construction process, we mitigated the class imbalance problem. In a class-imbalance dataset, certain classes have very high or lower representation compared to others. iii) We then investigated the performance of DL models following several pipelines, e.g. transfer learning and fine-tuning. Finally, we provide a thorough analysis and offer future perspectives (Section \ref{result_discussion}).\par

The main contributions of this research are:

\begin{itemize}
    \item construction of a large data set by combining four small-scale datasets with a variety of weeds and crops.
    \item addressing the class imbalance issue of the combined dataset using the data augmentation technique.
    \item comparing the performance of five well-known DL methods using the combined dataset.
    \item evaluating the efficiency of the pre-trained models on the combined dataset using the transfer learning and fine-tuning approach.
\end{itemize} 

This paper is organised as follows:  Section \ref{material_method} describes the materials and methods, including datasets, pre-processing approaches of images, data augmentation techniques, DL architectures and performance metrics. Section \ref{result_discussion} covers the experimental results and analysis, and Section \ref{conclusion} concludes the paper. \par

\section{Materials and Methods} \label{material_method}

\subsection{Dataset} 

In this work, four publicly available datasets were used: the \enquote{DeepWeeds dataset}\parencite{olsen2019deepweeds}, the \enquote{Soybean weed dataset} \parencite{dos2017weed}, the \enquote{Cotton-tomato and weed dataset} \parencite{espejo2020towards} and the \enquote{Corn weed dataset} \parencite{jiang2020cnn}. 

\subsubsection{DeepWeeds dataset}
The DeepWeeds dataset contains images of eight nationally significant species of weeds collected from eight rangeland environments across northern Australia. It also includes another class of images that contain non-weed plants. These are represented as a negative class. In this research, the negative image class was not used as it does not have any weed species. The images were collected using a FLIR Blackfly 23S6C high-resolution (1920 $\times$ 1200 pixel) camera paired with the Fujinon CF25HA-1 machine vision lens \parencite{olsen2019deepweeds}. The dataset is publicly available through the GitHub repository: \url{https://github.com/AlexOlsen/DeepWeeds}. \par 

\subsubsection{Soybean Weed Dataset}
\textcite{dos2017weed} acquired soybean, broadleaf, grass and soil images from Campo Grande in Brazil. We did not use the images from the  soil class as they did not contain crop plants or weeds. \textcite{dos2017weed} used a \enquote{Sony EXMOR} RGB camera mounted on an Unmanned Aerial Vehicle (UAV - - DJI Phantom 3 Professional). The flights were undertaken in the morning (8 to 10 am) from December 2015 to March 2016 with 400 images captured manually at an average height of four meters above the ground. The images of size 4000 $\times$ 3000 were then segmented using the Simple Linear Iterative Clustering (SLIC) superpixels algorithm \parencite{achanta2012slic} with manual annotation of the segments to their respective classes. The dataset contained 15336 segments of four classes. This dataset is publicly available at the website: \url{https://data.mendeley.com/datasets/3fmjm7ncc6/2}. \par

\subsubsection{Cotton Tomato Weed Dataset}
This dataset was acquired from three different farms in Greece, covering the south-central, central and northern areas of Greece. The images were captured in the morning (8 to 10 am) from May 2019 to June 2019 to ensure similar light intensities. The images of size 2272 $\times$ 1704 were taken manually from about one-meter height using a Nikon D700 camera \parencite{espejo2020towards}. The dataset is available through the GitHub repository: \url{https://github.com/AUAgroup/early-crop-weed}. \par

\subsubsection{Corn Weed Dataset}
This dataset was taken from a corn field in China. A total of 6000 images were captured using a Canon PowerShot SX600 HS camera placed vertically above the crop. To avoid the influence of illumination variations from different backgrounds, the images were taken under various lighting conditions. The original images were large (3264 $\times$ 2448), and these were subsequently resized to a resolution of 800 $\times$ 600 \parencite{jiang2020cnn}. The dataset is available at the Github: \url{https://github.com/zhangchuanyin/weed-datasets/tree/master/corn\%20weed\%20datasets}. \par

\subsubsection{Our Combined Dataset} \label{ComDataset}
In this paper, we combine all these datasets to create a single large dataset with weed and crop images sourced from different weather and geographical zones. This has created extra variability and complexity in the dataset with a large number of classes. This is also an opportunity to test the DL models and show their efficacy in complex settings. We used this combined dataset to train the classification models. Table \ref{tab:summary_dataset} provides a summary of the dataset used. The combined dataset contains four types of crop plants and sixteen species of weeds. The combined dataset is highly class-imbalanced since 27\% of images are from the soybean crop, while only 0.2\% of images are from the cotton crop (Table \ref{tab:summary_dataset}).

\subsubsection{Unseen Test Dataset} \label{UnseenTestData}
Another set of data was collected from the \href{https://edenlibrary.ai/}{Eden Library} website (\url{https://edenlibrary.ai/}) for this research. The website contains some plant datasets for different research work that use artificial intelligence. The images were collected under field conditions. We used images of five different crop plants from the website namely: Chinese cabbage (142 images), grapevine (33 images), pepper (355 images), red cabbage (52 images) and zucchini (100 images). In addition, we also included 500 images of lettuce plants \parencite{jiang2020cnn} and 201 images of radish plants \parencite{lameski2017weed} in the combined dataset. This dataset was then used to evaluate the performance of the transfer learning approach. This experiment checks the reusability of the DL models in the case of a new dataset. \par

\begin{table}[tb!]
\centering
\caption{Summary of crop and weed datasets used in this research}
\label{tab:summary_dataset}
\begin{tabular}{|c|c|c|l|c|c|}
\hline
\textbf{Dataset} & \textbf{Location} & \multicolumn{2}{c|}{\textbf{Crop/weed species}} & \textbf{\begin{tabular}[c]{@{}c@{}}Number\\of images\end{tabular}} & \textbf{\begin{tabular}[c]{@{}c@{}}\% of images in\\the class in the \\ combined dataset\end{tabular}} \\ \hline
\multirow{8}{*}{DeepWeeds} & \multirow{8}{*}{Australia} & \multirow{8}{*}{Weed} & Chinee apple & 1126 & 4.17 \\ \cline{4-6}
 & & & Lantana & 1063 & 3.94 \\ \cline{4-6}
 & & & Parkinsonia & 1031 & 3.82 \\ \cline{4-6}
 & & & Parthenium & 1022 & 3.78 \\ \cline{4-6}
 & & & Prickly acacia & 1062 & 3.93 \\ \cline{4-6}
 & & & Rubber vine & 1009 & 3.74 \\ \cline{4-6}
 & & & Siam weed & 1074 & 3.98 \\ \cline{4-6}
 & & & snakeweed & 1016 & 3.76 \\ \hline
\multirow{3}{*}{Soybean Weed} & \multirow{3}{*}{Brazil} & Crop & Soybean & 7376 & 27.31     \\ \cline{3-6}
 & & \multirow{2}{*}{Weed} & Broadleaf & 1191 & 4.41 \\ \cline{4-6}
 & & & Grass & 3526 & 13.06 \\ \hline
\multirow{4}{*}{\begin{tabular}[c]{@{}c@{}}Cotton Tomato \\ Weed\end{tabular}} & \multirow{4}{*}{Greece} & \multirow{2}{*}{Crop} & Cotton & 54 & 0.20 \\ \cline{4-6}
 & & & Tomato & 201 & 0.74 \\ \cline{3-6}
 & & \multirow{2}{*}{Weed} & Black nightshade & 123 & 0.46 \\ \cline{4-6}
 & & & Velvet leaf & 130 & 0.48 \\ \hline
\multirow{5}{*}{Corn Weed} & \multirow{5}{*}{China} & Crop & Corn & 1200 & 4.44      \\ \cline{3-6}
 & & \multirow{4}{*}{Weed} & Blue Grass & 1200 & 4.44 \\ \cline{4-6}
 & & & Chenopodium album & 1200 & 4.44 \\ \cline{4-6}
 & & & Cirsium setosum & 1200 & 4.44 \\ \cline{4-6}
 & & & Sedge & 1200 & 4.44 \\ \hline
\end{tabular}
\end{table}

\begin{figure}[ht!]
    \centering
    \captionsetup[subfigure]{justification=centering}
    \begin{subfigure}[b]{.23\textwidth}
        \centering
        \includegraphics[width=\textwidth, height = 8em]{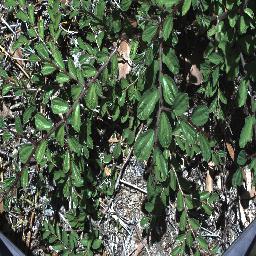}
        \caption{\scriptsize{Chinee apple}}
        \label{fig:ca}
    \end{subfigure}
    \vspace{1em}
    \begin{subfigure}[b]{.23\textwidth}
        \centering
        \includegraphics[width=\textwidth, height = 8em]{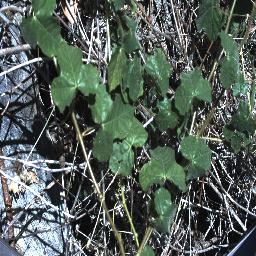}  
        \caption{\scriptsize{Lantana}}
        \label{fig:lan}
    \end{subfigure}
    \begin{subfigure}[b]{.23\textwidth}
        \centering
        \includegraphics[width=\textwidth, height = 8em]{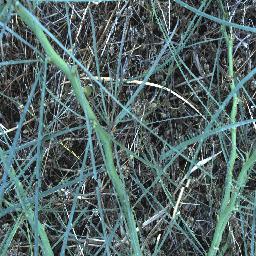}  
        \caption{\scriptsize{Parkinsonia}}
        \label{fig:park}
    \end{subfigure}
    \begin{subfigure}[b]{.23\textwidth}
        \centering
        \includegraphics[width=\textwidth, height = 8em]{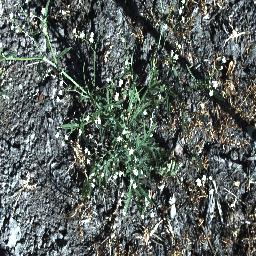}  
        \caption{\scriptsize{Parthenium}}
        \label{fig:parth}
    \end{subfigure}
    \vspace{1em}
    \begin{subfigure}[b]{.23\textwidth}
        \centering
        \includegraphics[width=\textwidth, height = 8em]{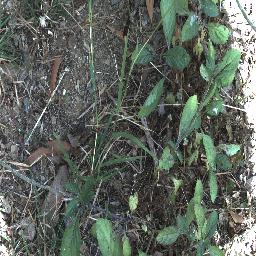}  
        \caption{\scriptsize{Prickly acacia}}
        \label{fig:prickly}
    \end{subfigure}
    \begin{subfigure}[b]{.23\textwidth}
        \centering
        \includegraphics[width=\textwidth, height = 8em]{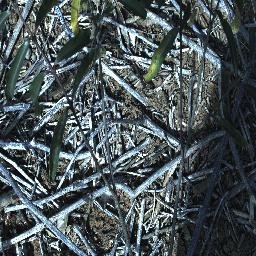}  
        \caption{\scriptsize{Rubber vine}}
        \label{fig:rubber}
    \end{subfigure}
    \begin{subfigure}[b]{.23\textwidth}
        \centering
        \includegraphics[width=\textwidth, height = 8em]{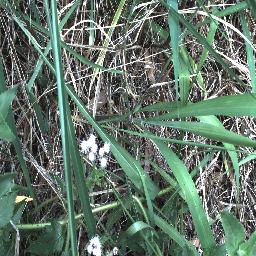}  
        \caption{\scriptsize{Siam weed}}
        \label{fig:siam}
    \end{subfigure}
    \begin{subfigure}[b]{.23\textwidth}
        \centering
        \includegraphics[width=\textwidth, height = 8em]{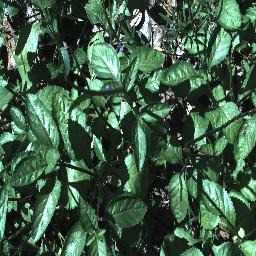}  
        \caption{\scriptsize{snakeweed}}
        \label{fig:snake}
    \end{subfigure}
    \vspace{1em}
    \begin{subfigure}[b]{.23\textwidth}
        \centering
        \includegraphics[width=\textwidth, height = 8em]{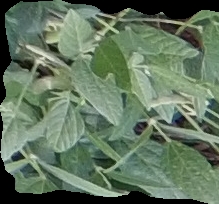}  
        \caption{\scriptsize{Soybean}}
        \label{fig:soybean}
    \end{subfigure}
    \begin{subfigure}[b]{.23\textwidth}
        \centering
        \includegraphics[width=\textwidth, height = 8em]{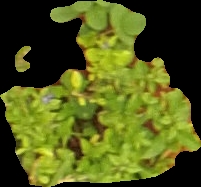}  
        \caption{\scriptsize{Broadleaf}}
        \label{fig:Broadleaf}
    \end{subfigure}
    \begin{subfigure}[b]{.23\textwidth}
        \centering
        \includegraphics[width=\textwidth, height = 8em]{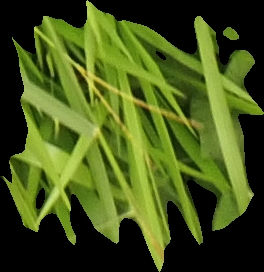}  
        \caption{\scriptsize{Grass}}
        \label{fig:Grass}
    \end{subfigure}
    \begin{subfigure}[b]{.23\textwidth}
        \centering
        \includegraphics[width=\textwidth, height = 8em]{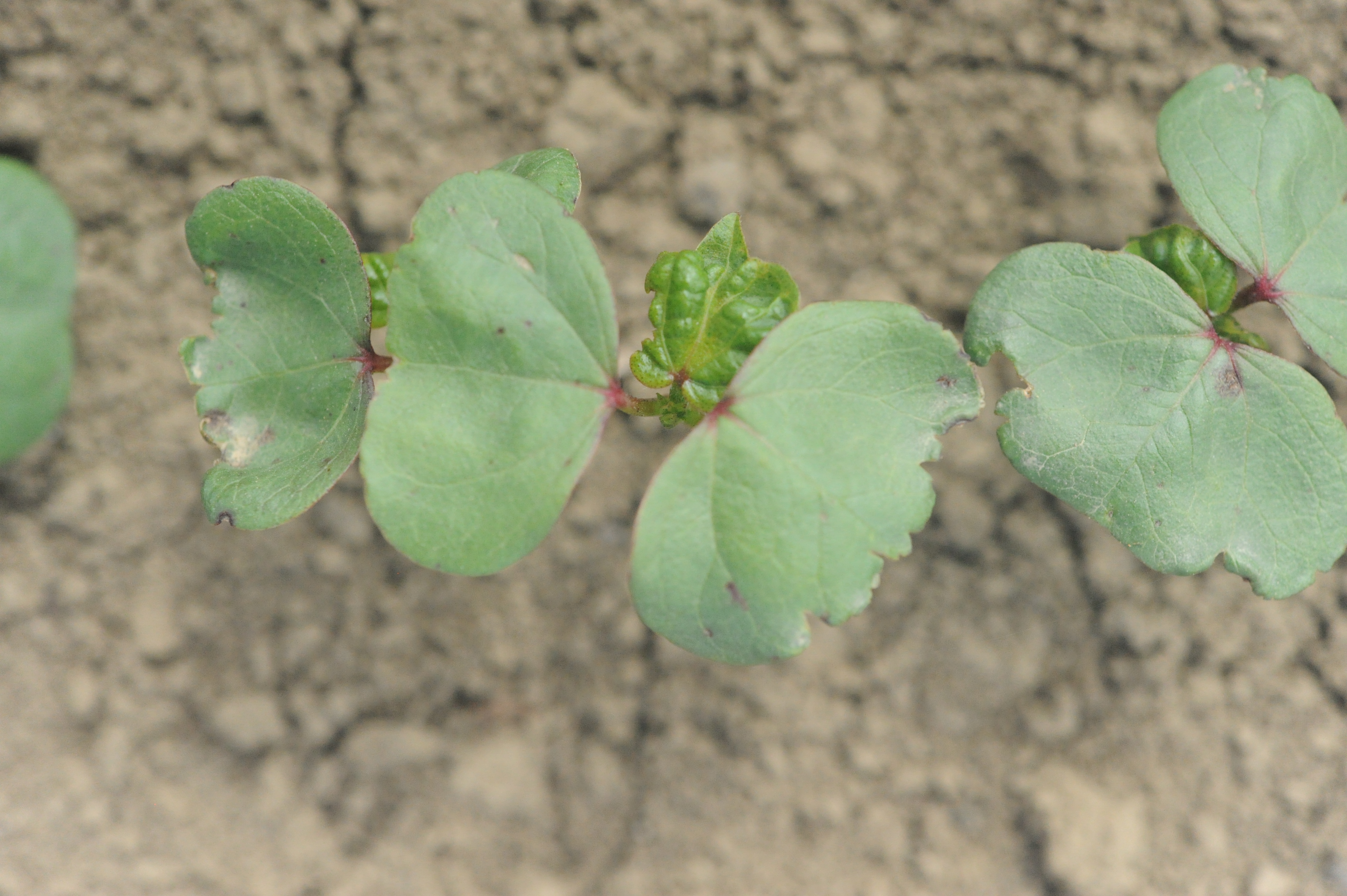}  
        \caption{\scriptsize{Cotton}}
        \label{fig:Cotton}
    \end{subfigure}
    \vspace{1em}
    \begin{subfigure}[b]{.23\textwidth}
        \centering
        \includegraphics[width=\textwidth, height = 8em]{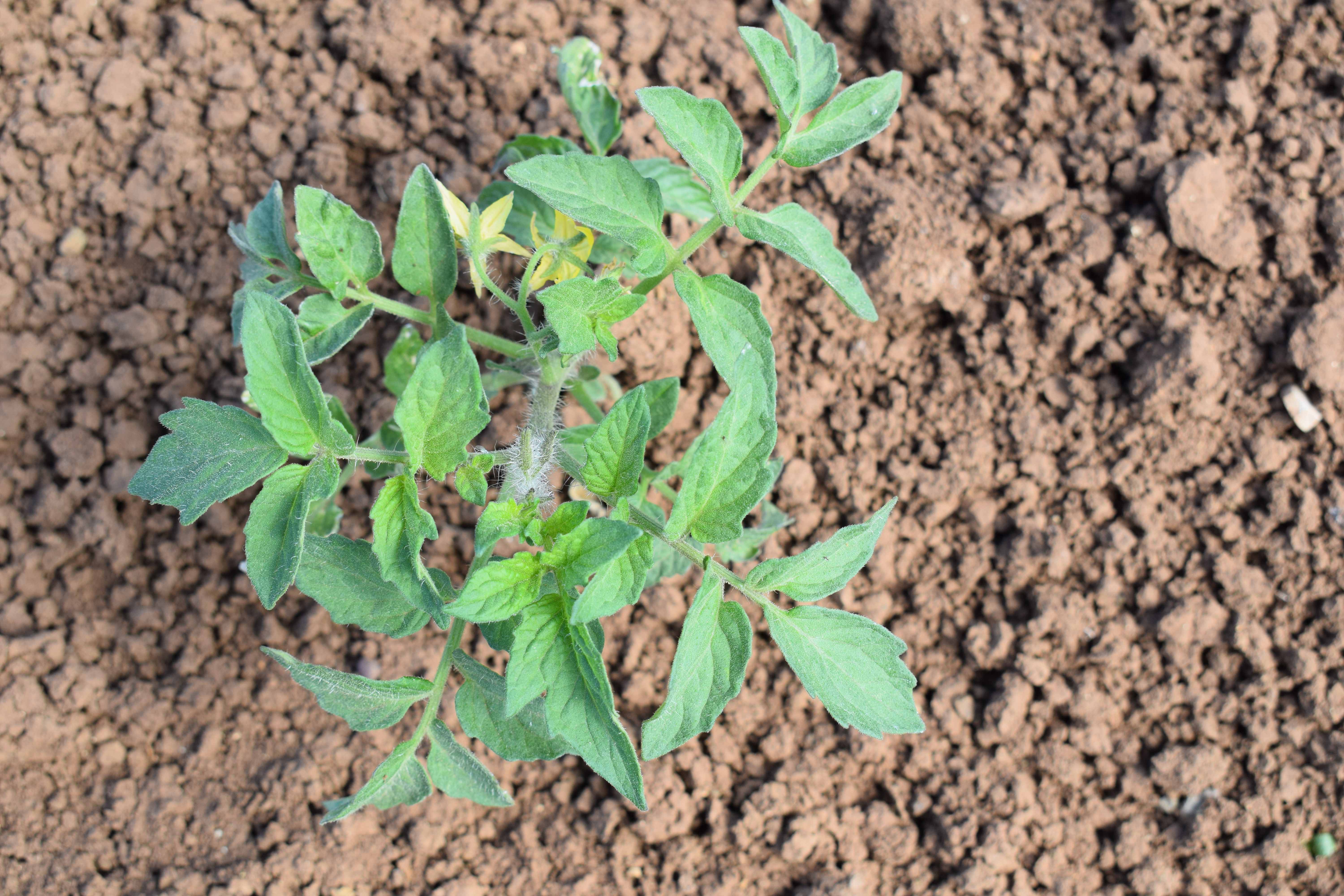}  
        \caption{\scriptsize{Tomato}}
        \label{fig:Tomato}
    \end{subfigure}
    \begin{subfigure}[b]{.23\textwidth}
        \centering
        \includegraphics[width=\textwidth, height = 8em]{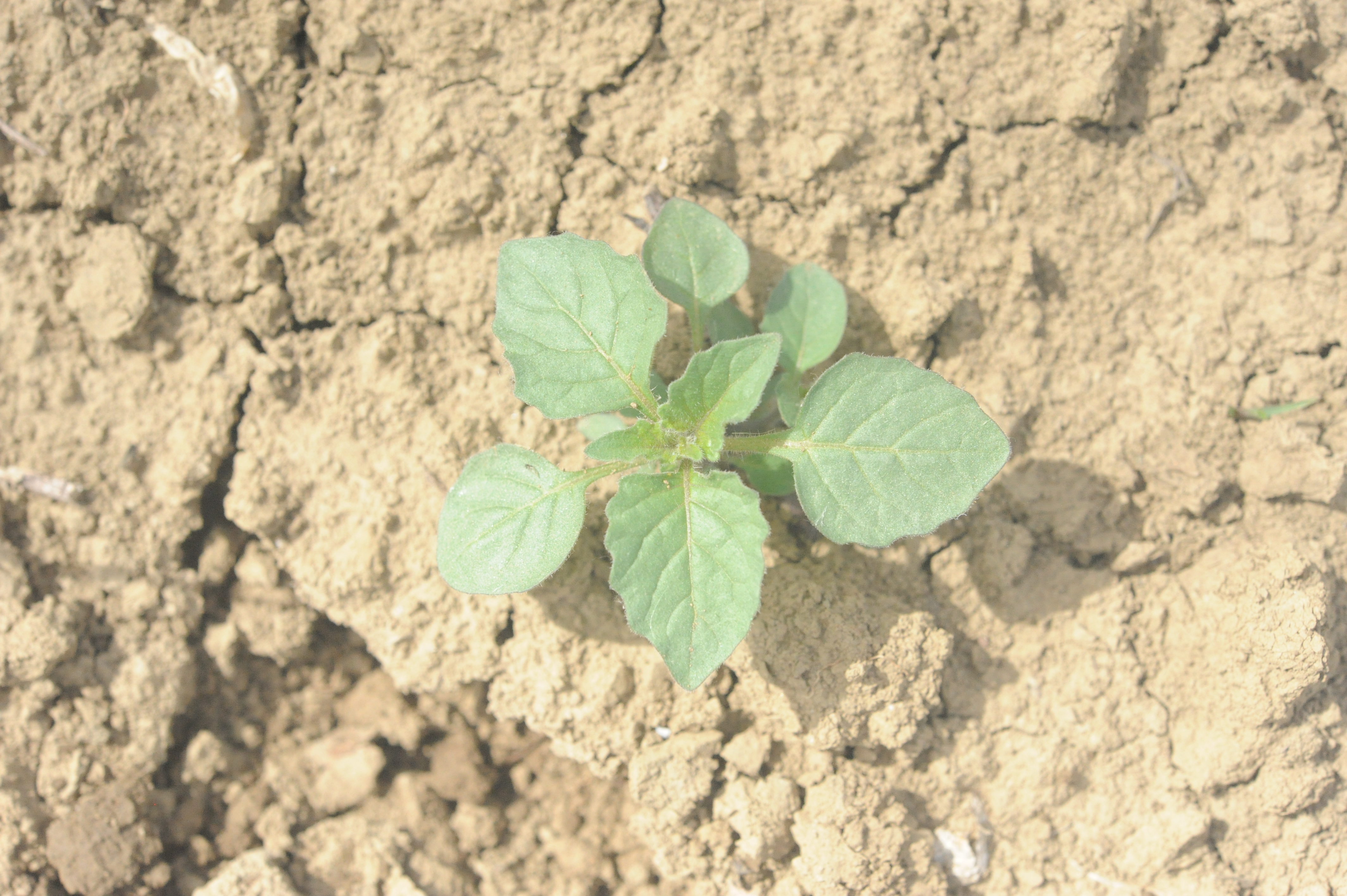}  
        \caption{\scriptsize{Black nightshade}}
        \label{fig:Black}
    \end{subfigure}
    \begin{subfigure}[b]{.23\textwidth}
        \centering
        \includegraphics[width=\textwidth, height = 8em]{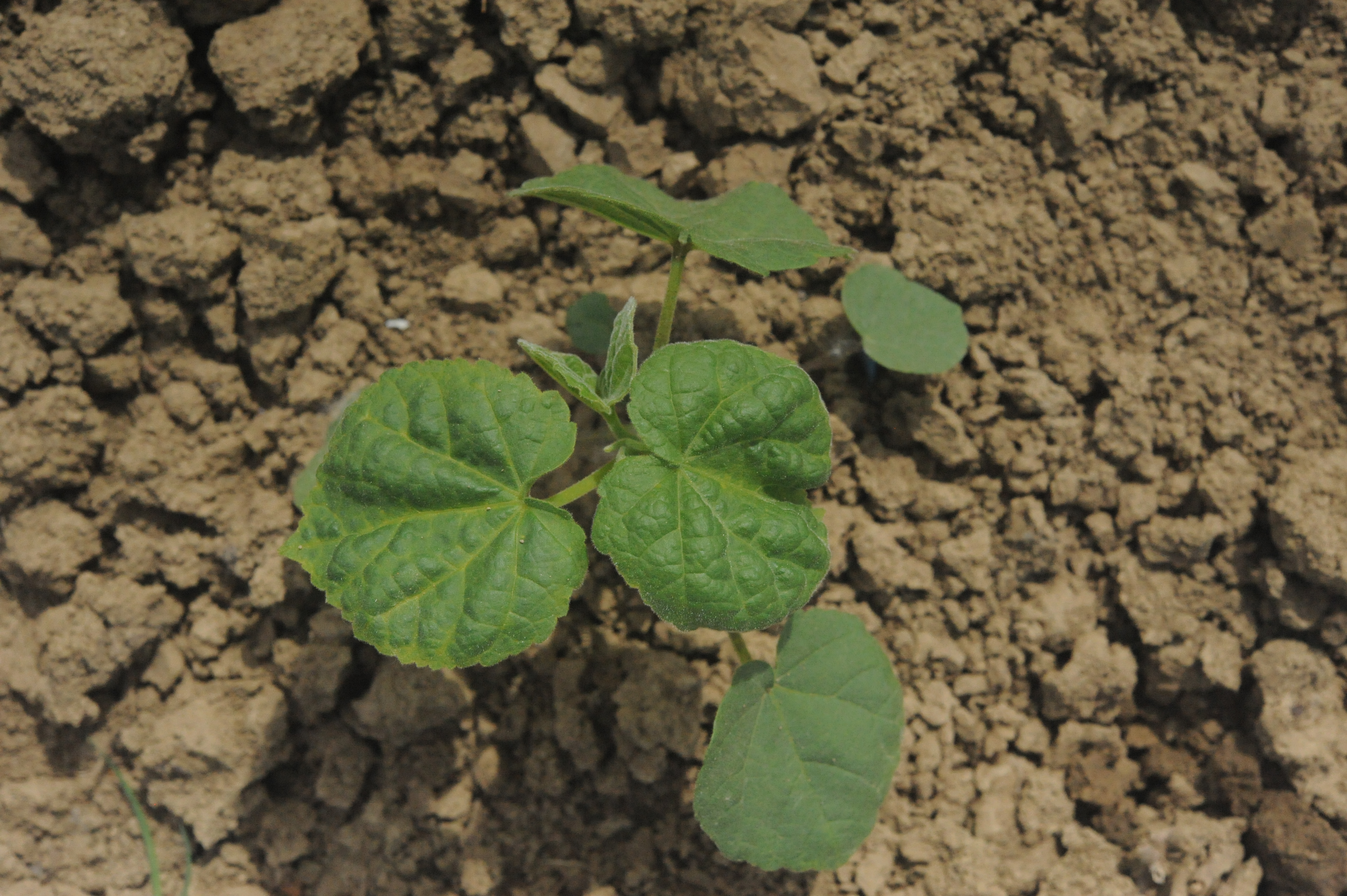}  
        \caption{\scriptsize{Velvet leaf}}
        \label{fig:Velvet}
    \end{subfigure}
    \begin{subfigure}[b]{.23\textwidth}
        \centering
        \includegraphics[width=\textwidth, height = 8em]{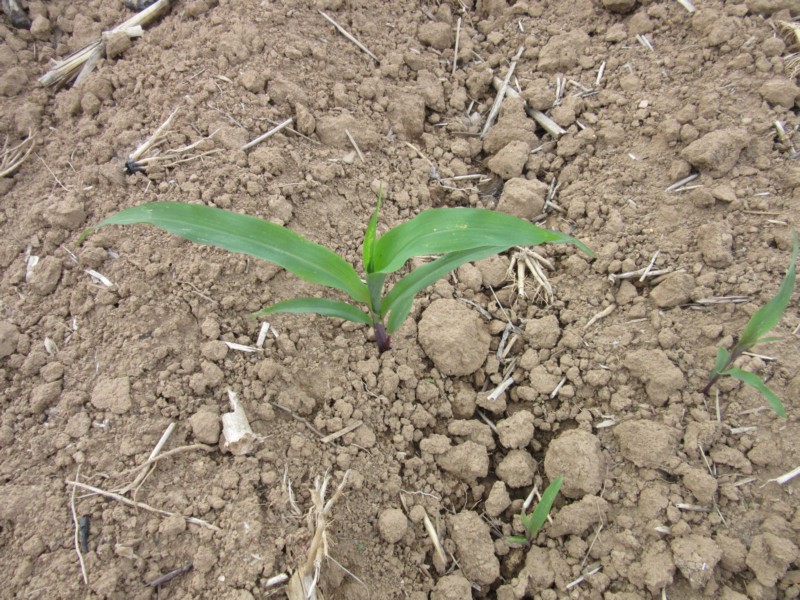}  
        \caption{\scriptsize{Corn}}
        \label{fig:corn}
    \end{subfigure}
    \vspace{1em}
    \begin{subfigure}[b]{.23\textwidth}
        \centering
        \includegraphics[width=\textwidth, height = 8em]{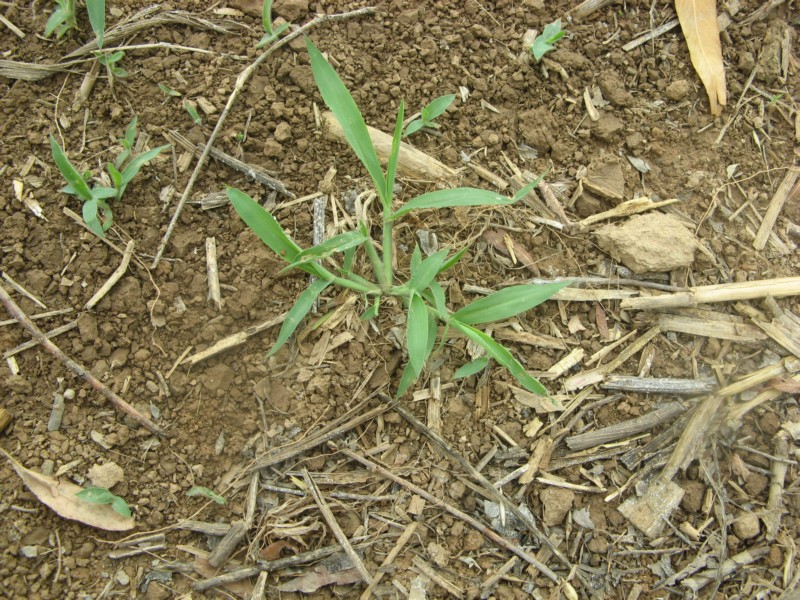}  
        \caption{\scriptsize{Blue grass}}
        \label{fig:Blue}
    \end{subfigure}
    \begin{subfigure}[b]{.23\textwidth}
        \centering
        \includegraphics[width=\textwidth, height = 8em]{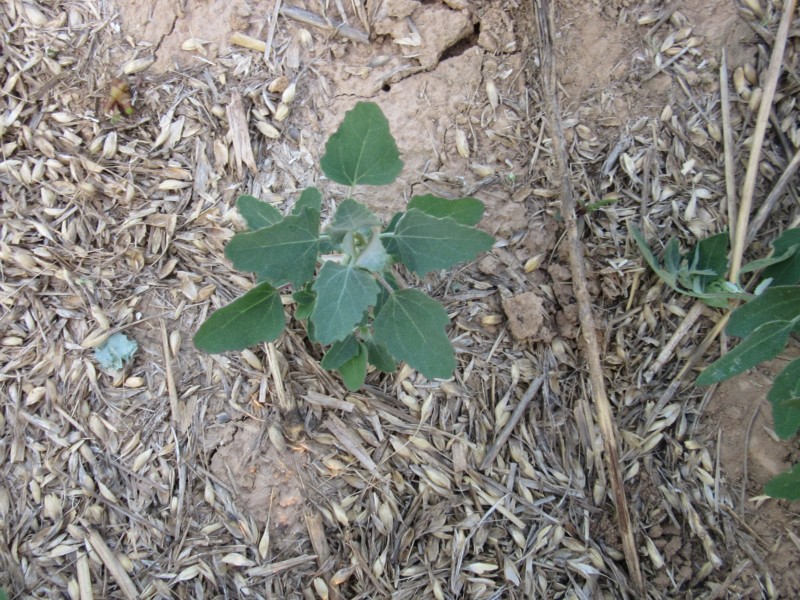}  
        \caption{\scriptsize{Chenopodium album}}
        \label{fig:Chenopodium}
    \end{subfigure}
    \begin{subfigure}[b]{.23\textwidth}
        \centering
        \includegraphics[width=\textwidth, height = 8em]{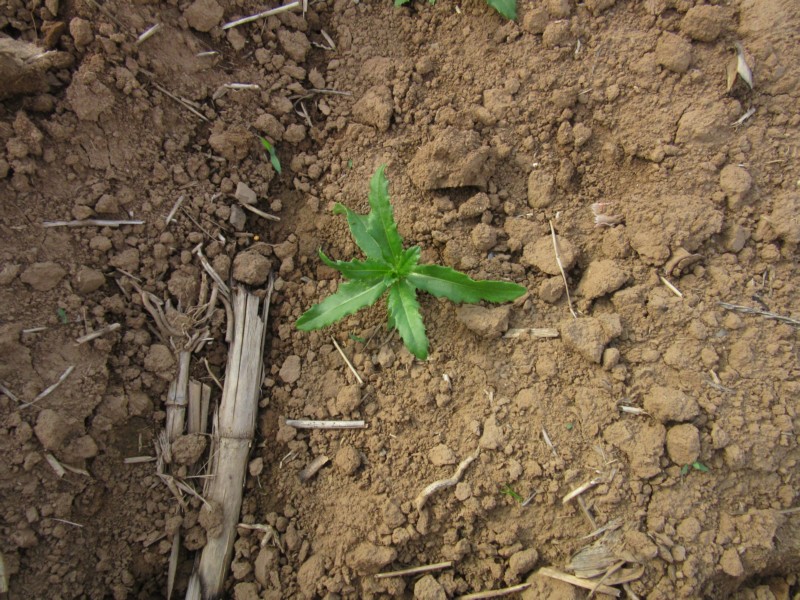}  
        \caption{\scriptsize{Cirsium setosum}}
        \label{fig:Cirsium}
    \end{subfigure}
    \begin{subfigure}[b]{.23\textwidth}
        \centering
        \includegraphics[width=\textwidth, height = 8em]{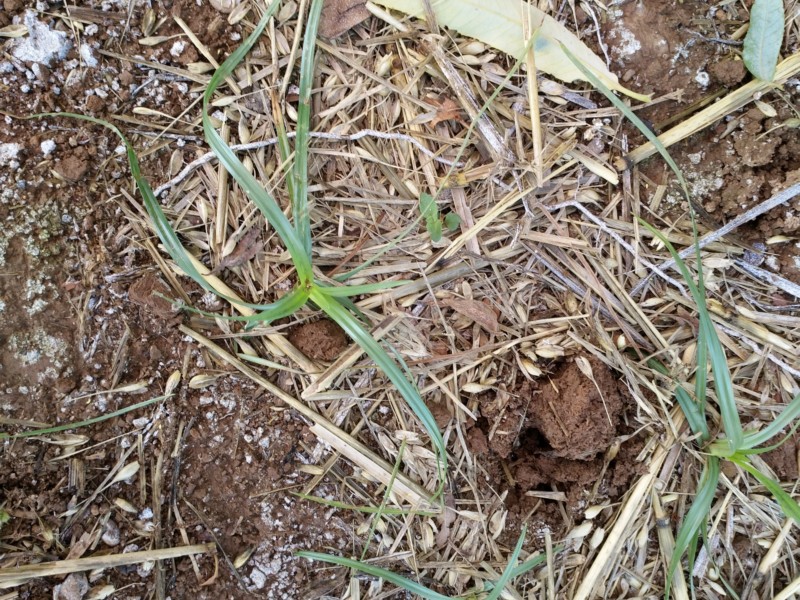}  
        \caption{\scriptsize{Sedge}}
        \label{fig:Sedge}
    \end{subfigure}

\caption[Sample crop and weed images of each class from the datasets.]{Sample crop and weed images of each class from the datasets.}
\label{fig:sample_crop_weed}
\end{figure}

In the study, the images of each class were randomly assigned for training (60\%), validation (20\%) and testing (20\%). Each image was labelled with one image-level annotation. This means that each image has only one label, i.e., the name of the weed or crop classes, e.g., Chinee apple or corn. Figure \ref{fig:sample_crop_weed} provides sample images in the dataset.

\subsection{Image Pre-processing}
Some level of image pre-processing is needed before the data can be used as input for training the DL model. This includes resizing the images, removing the background, enhancing and denoising the images, colour transformation, morphological transformation etc. In this study, the Keras pre-processing utilities \parencite{chollet2015keras} were used to prepare the data for training. This function applies some predefined operations to the data. One of the operations is to increase the dimension of the input. DL models process images in batches. To create the batches of images, additional dimension resizing is needed. An image contains three properties, e.g., image height, width and the number of channels. The pre-processing function adds a dimension to the image for inclusion in the batch information. Pre-processing involves normalising the data so that the pixel values range is from 0 to 1. Each model has a specific pre-processing technique to transform a standard image into an appropriate input. Research suggests that the classification model performance is improved by increasing the input resolution of the images \parencite{sabottke2020effect,sahlsten2019deep}. However, the model's computational complexity also increases with a higher resolution of the input image. The default input resolution for all the models used in this research is 224 $\times$ 224. \par 

\subsection{Data Augmentation}
The combined dataset is highly class-imbalanced. The minority classes are over-sampled using image augmentation to balance the dataset. The augmented data is only used to train the models. Image augmentation is done using the Python image processing library Scikit-image \parencite{van2014scikit}. After splitting the dataset into training, validation and testing sets, most training images were from soybean with 4,425 image. By applying augmentation approaches, we obtained 4425 images for all other weed and crop classes; thus we ensured that all classes were balanced. The following operations were applied randomly to the data to generate the augmented images:

\begin{itemize}
    \item Random rotation in the range of [-25, +25] degrees,
    \item Horizontal and vertical scaling in the range of 0.5 and 1, 
    \item Horizontal and vertical flip,
    \item Added random noise (Gaussian noise),
    \item Blurring the images,
    \item Applied gamma, sigmoid and logarithmic correction operation, and
    \item Stretched or shrunk the intensity levels of images.
\end{itemize}

The models are then trained on both actual data and augmented data without making any discrimination. \par

\subsection{Deep Learning} \label{DL}
Five state-of-the-art deep learning models with pre-trained weights were used in this research to classify images. These models were made available via the Keras Application Programming Interface (API) \parencite{chollet2015keras}. TensorFlow \parencite{abadi2016tensorflow} was used as a machine learning framework. The selected CNN architectures were:

\begin{itemize}
    \item \textbf{VGG16} \parencite{simonyan2014very} uses a stack of convolutional layers with a very small receptive field (3 $\times$ 3). It was the winner of ImageNet Challenge 2014 in the localisation track. The architecture consists of a stack of 13 convolutional layers, followed by three fully connected layers. A very small receptive field (3 $\times$ 3) is used in the convolutional layers. The network fixes the convolutional stride and padding to 1 pixel. Spatial pooling is carried out by the max-pooling layers. However, only five of the convolutional layers are followed by the max-pooling layer. This actual state-of-the-art VGG16 model has 138,357,544 trainable parameters. Of these, about 124 million parameters are contained in the fully connected layers. Those layers were customised in this research.  
    
    \item \textbf{ResNet-50} \parencite{he2016deep} is deeper than VGG16 but has a lower computational complexity. Generally, with increasing depths of the network, the performance becomes saturated or degraded. The model uses residual blocks to maintain accuracy with the deeper network. The residual blocks also contain convolutions layers like VGG16. The model uses batch normalisation after each convolutional layer and before the activation layer. The model explicitly reformulates the layers as residual functions with reference to the input layers and skip connections. Although the model contains more layers than VGG16, it only has 25,636,712 trainable parameters. 
    
    \item \textbf{Inception-V3} \parencite{szegedy2016rethinking} uses a deeper network with fewer training parameters (23,851,784). The model consists of symmetric and asymmetric building blocks with convolutions, average pooling, max pooling, concats, dropouts, and fully connected layers.
    
    \item \textbf{Inception-ResNet-V2} \parencite{szegedy2017inception} combines the concept of skip connections from ResNet with Inception modules. Each inception block is followed by a filter expansion layer (1 $\times$ 1 convolution without activation). Before concatenation with the input layer the dimensionality expansion is performed to match the depth. The model uses batch normalisation only on the traditional layer, but not for the summation layers. The network is 164 layers deep and has 55,873,736 trainable parameters.
    
    \item  MobileNetV2 \parencite{sandler2018mobilenetv2} allows memory-efficient inference with a reduced number of parameters. It contains 3,538,984 trainable parameters. The basic building block of the model is a bottleneck depth-separable convolution with residuals. The model has the initial fully convolution layer with 32 filters, followed by 19 residual bottleneck layers. It always uses 3 $\times$ 3 kernels and utilises the dropout layer and batch normalisation during training. Instead of ReLU (Rectified Linear Unit), this model uses ReLU6 as an activation function. ReLU6 is a variant of ReLU, where the number 6 is an arbitrary choice of the upper bound, which worked well and the model can easily learn the sparse features.
\end{itemize}

All the models were initialised with pre-trained weights trained on the ImageNet dataset. As the models were trained to recognise 1000 different objects, the original architecture was slightly modified to classify twenty crops and weed species. The last fully-connected layer of the original model was replaced by a global average pooling layer followed by two dense layers with 1024 neurons and \enquote{ReLU} activation function. The output contained another dense layer where the number of neurons depended on the number of classes. The softmax activation function was used in the output layer since the models were multi-class classifiers. The size of the input was 256$\times$256$\times$3, and the batch size was 64. The maximum number of epochs for training the models was 100. However, often the training was completed before reaching the maximum number. The initial learning rate was set to $1 \times 10$\textsuperscript{-4} and is randomly decreased down to 10\textsuperscript{-6} by monitoring the validation loss in every epoch. Table \ref{tab:num_param} shows the number of parameters of each of the models used in this research without the output layer. It was found that the Inception-Resnet-V2 model has the most parameters, and the MobileNetV2 model has the least.

\begin{table}[tb]
    \centering
    \caption{Number of parameters used in the deep learning models}
    \label{tab:num_param}
    \begin{tabular}{|l|r|}\hline
      \textbf{Deep Learning Model} & \textbf{Number of parameters} \\ \hline \hline
        VGG16 & 16,289,600 \\ \hline
        ResNet-50 & 26,735,488 \\ \hline
        Inception-V3 & 24,950,560 \\ \hline
        Inception-ResNet-V2  & 56,960,224 \\ \hline
        MobileNetV2 & 4,585,216 \\ \hline
      \end{tabular}
\end{table}

\subsection{Transfer Learning and Fine-Tuning}

\begin{figure}[ht]
    \centering
    \captionsetup[subfigure]{justification=centering}
    \begin{subfigure}[b]{0.95\textwidth}
        \centering
        \includegraphics[width=\textwidth, height = 10em]{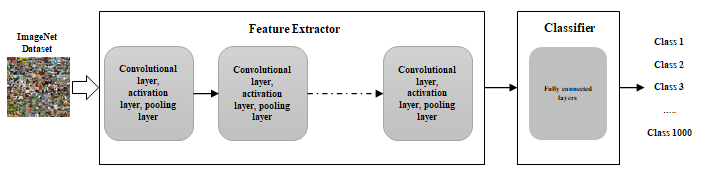}
        \caption{\scriptsize{Pre-trained DL model}}
        \label{fig:sota}
    \end{subfigure}
    \vspace{1em}
    \begin{subfigure}[b]{.95\textwidth}
        \centering
        \includegraphics[width=\textwidth, height = 13em]{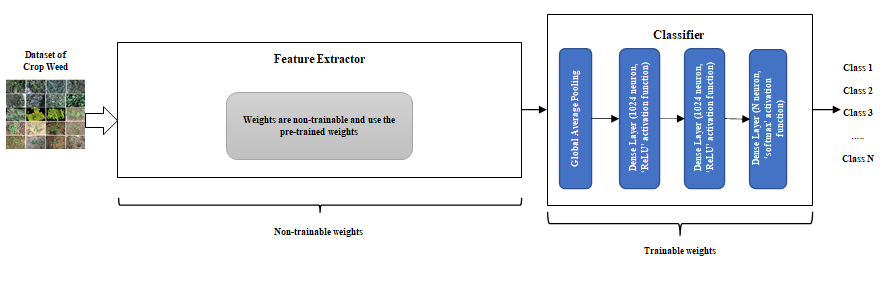}  
        \caption{\scriptsize{Transfer learning approach}}
        \label{fig:tl_app}
    \end{subfigure}
    \begin{subfigure}[b]{.95\textwidth}
        \centering
        \includegraphics[width=\textwidth, height = 13em]{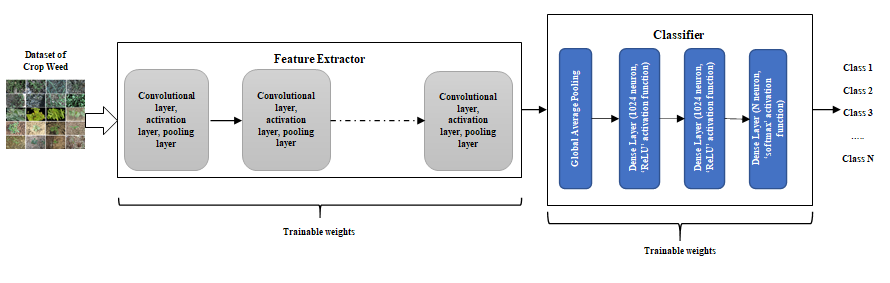}  
        \caption{\scriptsize{Fine-tuning approach}}
        \label{fig:ft_app}
    \end{subfigure}
\caption[The basic block diagram of DL models used for the experiments.]{The basic block diagram of DL models used for the experiments.}
\label{fig:arc}
\end{figure}

A conventional DL model contains two basic components: a feature extractor and a classifier. Depending on the DL model, different layers in the feature extractor and classifier may vary. However, all the DL architectures, used in this research, contain a series of trainable filters. Their weights are adjusted or trained for classifying images of a target dataset. Figure \ref{fig:sota} shows a basic structure of a pre-trained DL model. A pre-trained DL model means that the weights of the filters in the feature extractor and classifier is trained to classify 1000 different classes of images contained in the ImageNet dataset. The concept of transfer learning is to use those pre-trained weights to classify the images of a new unseen dataset \parencite{guo2019spottune, pan2009survey}. We used this approach in two different ways. The approaches were categorised as transfer learning and fine-tuning. To train the model using our dataset of crop and weed images, we took the feature extractor from the pre-trained DL model and removed its classifier part since it was designed for a specific classification task. In the transfer learning approach (Figure \ref{fig:tl_app}), we only trained the weights of the filters in the classifier part and kept the pre-trained weights of the layer in the feature extractor. This process eliminates the potential issue of training the complete network on a large number of labelled images. However, in the fine-tuning approach (Figure \ref{fig:ft_app}), the weights in the feature extractor were initialised from the pre-trained model, but not fixed. During the training phase of the model, the weights were retrained together with the classifier part. This process increased the  efficiency of the classifier because it was not necessary to train the whole model from scratch. The model can extract discriminating features for the target dataset more accurately. Our experiments used both approaches and evaluated their performance on the crop and weed image dataset. Finally, we trained one state-of-the-art DL architecture from scratch, using our combined dataset (Section \ref{ComDataset}) and used its feature extractor to classify the images in an unseen test dataset (Section \ref{UnseenTestData}) using the transfer learning approach. The performance of the pre-trained state-of-the-art model was then compared with the model trained on the crop and weed dataset. 

\subsection{Performance Metrics}
The models were tested and thoroughly evaluated using several metrics: accuracy, precision, recall, and F1 score metrics, which are defined as follows:

\begin{itemize}
    \item \textbf{Accuracy (Acc):} It is the percentage of images whose classes are predicted correctly among all the test images. A higher value represents a better result.
    \item \textbf{Precision (P):} The fraction of correct prediction (True Positive) from the total number of relevant result (Sum of True Positives and False Positives). 
    \item \textbf{Recall (R):} The fraction of True Positive from the sum of True Positive and False Negative (number of incorrect predictions).
    \item \textbf{F1 Score (F1):} The harmonic mean of precision and recall. This metric is useful to measure the performance of a model on a class-imbalanced dataset.
    \item \textbf{Confusion Matrix:} It is used to measure the performance of machine learning models for classification problems. The confusion matrix tabulates the comparison of the actual target values with the values predicted by the trained model. It helps to visualise how well the classification model is performing and what prediction errors it is making.
\end{itemize}

In all these metrics, a higher value represents better performance.

\section{Results and Discussions} \label{result_discussion}

We conducted five sets of experiments on the data. Table \ref{tab:num_dtvt} shows the number of images used for training, validation and testing of the models. Augmentation was applied to generate 4,425 images for each of the classes. However, only actual images were used to validate and test the models. All the experiments were done on a desktop computer, with an Intel(R) Core(TM) i9-9900X processor, 128 gigabyte of RAM and a NVIDIA GeForce RTX 2080 Ti Graphics Processing Unit (GPU). We used the Professional Edition of the Windows 10 operating system. The deep learning models were developed using Python 3.8 and Tensorflow 2.4 framework. \par

\begin{table}[tb]
\centering
\caption{The numbers of images used to train (after augmentation), validate and test the models.}
\label{tab:num_dtvt}
\small
\begin{tabular}{|p{0.1\textwidth}|p{0.15\textwidth}|M{0.15\textwidth}|M{0.15\textwidth}|M{0.13\textwidth}|M{0.12\textwidth}|}
\hline
\multirow{2}{*}{\textbf{Dataset}} & 
\multirow{2}{*}{\textbf{\begin{tabular}[c]{@{}c@{}}Crop and weed \\species\end{tabular}}} &
\multicolumn{2}{c|}{\textbf{Training set (number of images)}} &
\multirow{2}{*}{\textbf{\begin{tabular}[c]{@{}c@{}}Validation set \\(number of \\images)\end{tabular}}} &
\multirow{2}{*}{\textbf{\begin{tabular}[c]{@{}c@{}}Test set \\(number of \\images)\end{tabular}}}\\ \cline{3-4}
& & \textbf{Real images} & \textbf{Real and augmented images} & & \\ \hline
\multirow{8}{*}{DeepWeeds} & Chinee apple & 675 & 4,425 & 225 & 226 \\ \cline{2-6}
& Lantana & 637 & 4,425 & 212 & 214 \\ \cline{2-6}
& Parkinsonia & 618 & 4,425 & 206 & 207 \\ \cline{2-6}
& Parthenium & 613 & 4,425 & 204 & 205 \\ \cline{2-6}
& Prickly acacia & 637 & 4,425 & 212 & 213 \\ \cline{2-6}
& Rubber vine & 605 & 4,425 & 201 & 203 \\ \cline{2-6}
& Siam weed & 644 & 4,425 & 214 & 216 \\ \cline{2-6}
& snakeweed & 609 & 4,425 & 203 & 204 \\ \hline
\multirow{3}{*}{\begin{tabular}[l]{@{}l@{}}Soybean \\Weed\end{tabular}} & Soybean & 4,425 & 4,425 & 1,475 & 1,476 \\ \cline{2-6}
& Broadleaf & 714 & 4,425 & 238 & 239 \\ \cline{2-6}
& Grass & 2,112 & 4,425 & 704 & 704 \\ \hline
\multirow{4}{*}{\begin{tabular}[l]{@{}l@{}}Cotton\\Tomato \\Weed\end{tabular}} & Cotton & 32 & 4,425 & 10 & 12 \\ \cline{2-6}
& Tomato & 120 & 4,425 & 40 & 41 \\ \cline{2-6}
& Black nightsade & 73 & 4,425 & 24 & 26 \\ \cline{2-6}
& Velvet leaf & 78 & 4,425 & 26 & 26 \\ \hline
\multirow{5}{*}{\begin{tabular}[l]{@{}l@{}}Corn \\Weed\end{tabular}} & Corn & 720 & 4,425 & 240 & 240 \\ \cline{2-6}
& Bluegrass & 720 & 4,425 & 240 & 240 \\ \cline{2-6}
& Chenopodium album & 720 & 4,425 & 240 & 240 \\ \cline{2-6}
& Cirsium setosum & 720 & 4,425 & 240 & 240 \\ \cline{2-6}
& Sedge & 718 & 4,425 & 239 & 241 \\ \hline
\end{tabular}
\end{table}

\subsection{Experiment 1: Comparing the performance of DL models for classifying images in each of the datasets}

In this experiment, we trained the five models separately on each dataset using only actual images (see Table \ref{tab:num_dtvt}). Both transfer learning (TL) and fine-tuning (FT) approaches were used to train the models. Table \ref{tab:TVTaccuracy} shows the training, validation and testing accuracy for the five models. \par

\begin{table}[]
\begin{center}
\caption{Training, validation and testing accuracy for classifying crop and weed species of all four datasets using different DL models.}
\label{tab:TVTaccuracy}
\begin{tabular}{|l|l|r|r|r|r|r|r|}
\hline
\multicolumn{1}{|c|}{\multirow{2}{*}{\textbf{Dataset}}} & \multicolumn{1}{c|}{\multirow{2}{*}{\textbf{\begin{tabular}[c]{@{}c@{}}Deep Learning \\ model\end{tabular}}}} & \multicolumn{2}{c|}{\textbf{\begin{tabular}[c]{@{}c@{}}Training \\ Accuracy (\%)\end{tabular}}} & \multicolumn{2}{c|}{\textbf{\begin{tabular}[c]{@{}c@{}}Validation \\ Accuracy (\%)\end{tabular}}} & \multicolumn{2}{c|}{\textbf{\begin{tabular}[c]{@{}c@{}}Testing \\ Accuracy (\%)\end{tabular}}} \\ \cline{3-8} 
\multicolumn{1}{|c|}{} & \multicolumn{1}{c|}{} & \multicolumn{1}{c|}{\textbf{TL}} & \multicolumn{1}{c|}{\textbf{FT}} & \multicolumn{1}{c|}{\textbf{TL}} & \multicolumn{1}{c|}{\textbf{FT}} & \multicolumn{1}{c|}{\textbf{TL}} & \multicolumn{1}{c|}{\textbf{FT}} \\ \hline

\multirow{5}{*}{DeepWeeds} & VGG16 & \textbf{98.43} & 99.46 & \textbf{83.84} & \textbf{93.44} & \textbf{84.05} & 93.36 \\ \cline{2-8} 
 & ResNet-50 & 97.56 & \textbf{100.00} & 46.51 & 92.96 & 44.31 & \textbf{93.78} \\ \cline{2-8} 
 & Inception-V3 & 81.20 & \textbf{100.00} & 34.28 & 86.17 & 34.77 & 86.08 \\ \cline{2-8} 
 & Inception-ResNet-V2 & 81.02 & \textbf{100.00} & 35.84 & 89.09 & 36.55 & 89.39 \\ \cline{2-8} 
 & MobileNetV2 & 96.47 & \textbf{100.00} & 35.01 & 33.09 & 32.23 & 31.87 \\ \hline\hline
\multirow{5}{*}{Corn Weed} & VGG16 & \textbf{100.00} & 99.97 & \textbf{96.83} & 99.33 & \textbf{96.92} & \textbf{99.67} \\ \cline{2-8} 
 & ResNet-50 & \textbf{100.00} & \textbf{100.00} & 71.72 & 99.50 & 63.11 & 99.50 \\ \cline{2-8} 
 & Inception-V3 & 98.92 & \textbf{100.00} & 68.39 & 98.41 & 59.28 & 98.42 \\ \cline{2-8} 
 & Inception-ResNet-V2 & 97.55 & \textbf{100.00} & 47.21 & \textbf{99.75} & 44.96 & 99.33 \\ \cline{2-8} 
 & MobileNetV2 & 99.03 & \textbf{100.00} & 70.89 & 89.91 & 69.03 & 87.51 \\ \hline\hline
\multirow{5}{*}{\begin{tabular}[c]{@{}l@{}}Cotton \\ Tomato \\ Weed\end{tabular}} & VGG16 & \textbf{100.00} & 96.04 & \textbf{94.00} & 92.00 & \textbf{99.05} & 88.57 \\ \cline{2-8} 
 & ResNet-50 & \textbf{100.00} & \textbf{100.00} & 54.00 & \textbf{99.00} & 55.24 & \textbf{99.05} \\ \cline{2-8} 
 & Inception-V3 & \textbf{100.00} & \textbf{100.00} & 53.00 & 96.00 & 59.05 & 98.10 \\ \cline{2-8} 
 & Inception-ResNet-V2 & 95.71 & \textbf{100.00} & 64.00 & 77.00 & 57.33 & 77.14 \\ \cline{2-8} 
 & MobileNetV2 & \textbf{100.00} & \textbf{100.00} & 64.00 & 72.00 & 60.00 & 78.10 \\ \hline\hline
\multirow{5}{*}{\begin{tabular}[c]{@{}l@{}}Soybean \\ Weed\end{tabular}} & VGG16 & \textbf{100.00} & 99.96 & \textbf{98.97} & 99.79 & \textbf{98.76} & \textbf{99.88} \\ \cline{2-8} 
 & ResNet-50 & 99.98 & \textbf{100.00} & 82.58 & \textbf{99.91} & 83.16 & 99.83 \\ \cline{2-8} 
 & Inception-V3 & 99.49 & \textbf{100.00} & 88.25 & 99.67 & 86.77 & 99.71 \\ \cline{2-8} 
 & Inception-ResNet-V2 & 98.80 & \textbf{100.00} & 90.36 & 99.79 & 89.78 & 99.59 \\ \cline{2-8} 
 & MobileNetV2 & \textbf{100.00} & \textbf{100.00} & 94.54 & 99.54 & 94.75 & 99.67 \\ \hline
\end{tabular}
\end{center}
\end{table}

On the DeepWeeds dataset, the VGG16 model achieved the highest training, validation and testing accuracy (98.43\%, 83.84\% and 84.05\% respectively) using the transfer learning approach. The training accuracy of the other four models was above 81\%. However, the validation and testing accuracy for those models were less than 50\%. This suggests that the models are overfitting. After fine-tuning the models, the overfitting problem was mitigated except for the MobileNetV2 architecture. Although four of the models achieved 100\% training accuracy after fine-tuning, the validation and testing accuracy was between 86\% and 94\%. MobileNetV2 model still overfitted even after fine-tuning with about 32\% validation and testing accuracy. Overall, the VGG16 model gave the best results for the DeepWeeds dataset as they had the least convolutional layers, which was adequate for small datasets. It should be noted that \textcite{olsen2019deepweeds}, who initially worked on this dataset, achieved an average classification accuracy of 95.1\% and 95.7\% using Inception-V3 and ResNet-50, respectively. However, they applied  data augmentation techniques to overcome the variable nature of the dataset.\par

On the Corn Weed and Cotton Tomato Weed datasets, the VGG16 and ResNet-50 models generally gave accurate result, but the accuracy of validation and testing were low for the DL models using the transfer learning approach for both datasets, and the classification performance of the models was substantially improved after fine-tuning. Among the five models, the retrained Inception-ResNet-V2 model gave better results for the Corn Weed dataset with training, validation and testing accuracy of 100\%, 99.75\% and 99.33\% respectively. The ResNet-50 model accurately classified the images of the Cotton Tomato Weed dataset.\par

VGG16 architecture reached about 99\% classification accuracy on both validation and testing data of the \enquote{Soybean Weed} dataset using the transfer learning approach. Also, the performance of four other models are better for this dataset using pre-trained weights. Compared to other datasets, the \enquote{Soybean Weed} dataset had more training samples, which helped to improve its classification performance. However, after fine-tuning the models on the datasets, all five deep learning architectures achieve more than 99\% classification accuracy on the validation and testing data.\par

According to the results of this experiment, as shown in Table \ref{tab:TVTaccuracy}, it can be concluded that, for classifying the images of crop and weed species dataset, the transfer learning approach does not work well. Since the pre-trained models were trained on the \enquote{ImageNet} dataset \parencite{deng2009imagenet}, which does not contain images of crop or weed species, the models cannot accurately classify weed images.\par

\subsection{Experiment 2: Combining two datasets} \label{comb_two}

In the previous experiment, we showed that it was unlikely to achieve better classification results using pre-trained weights for the convolutional layers of the DL models. The image classification accuracy improved by fine-tuning the weights of the models for the crop and weed dataset. For that reason, in this experiment, all the models were initialised with pre-trained weights and then retrained for the dataset. In this experiment, the datasets were paired up and used to generate six combinations to train the models. The training, validation and testing accuracies are shown in Table \ref{tab:two_dataset_com}. The combinations were-
\begin{itemize}
    \item \enquote{DeepWeeds} with \enquote{Corn Weed} dataset (DW-CW),
    \item \enquote{DeepWeeds} with \enquote{Cotton Tomato Weed} dataset (DW-CTW),
    \item \enquote{DeepWeeds} with \enquote{Soybean Weed} dataset (DW-SW),
    \item \enquote{Corn Weed} with \enquote{Cotton Tomato Weed} dataset (CW-CTW),
    \item \enquote{Corn Weed} with \enquote{Soybean Weed} dataset (CW-SW) and
    \item \enquote{Cotton Tomato Weed} with \enquote{Soybean Weed} dataset.
\end{itemize}

After fine-tuning the weights, all the DL models reached 100\% training accuracy. The accuracy of the DL architectures also gave better validation and testing results when trained with CW-CTW, CW-SW, CTW-SW combined datasets. However, the models overfitted when trained on the \enquote{DeepWeeds} dataset and combined with any of the other three datasets. \par

\begin{table}[]
\centering
\caption{Training, validation and testing accuracy of the DL models after training by combining two of the datasets}
\label{tab:two_dataset_com}
\begin{tabular}{|l|l|r|r|r|r|r|r|}
\hline
\multicolumn{1}{|c|}{\textbf{DL Models}}                                         & \multicolumn{1}{c|}{\textbf{Accuracy}} & \multicolumn{1}{c|}{\textbf{\begin{tabular}[c]{@{}c@{}}DW-\\ CW\end{tabular}}} & \multicolumn{1}{c|}{\textbf{\begin{tabular}[c]{@{}c@{}}DW-\\ CTW\end{tabular}}} & \multicolumn{1}{c|}{\textbf{\begin{tabular}[c]{@{}c@{}}DW-\\ SW\end{tabular}}} & \multicolumn{1}{c|}{\textbf{\begin{tabular}[c]{@{}c@{}}CW-\\ CTW\end{tabular}}} & \multicolumn{1}{c|}{\textbf{\begin{tabular}[c]{@{}c@{}}CW-\\ SW\end{tabular}}} & \multicolumn{1}{c|}{\textbf{\begin{tabular}[c]{@{}c@{}}CTW-\\ SW\end{tabular}}} \\ \hline\hline

 & Training& \textbf{100.00} & 99.63 & 99.95 & 99.97 & \textbf{100.00} & \textbf{100.00} \\ \cline{2-8} 
 & Validation & \textbf{96.21} & \textbf{93.64} & 97.31 & 98.99 & \textbf{99.67} & \textbf{99.76} \\ \cline{2-8} 

\multirow{-3}{*}{VGG16} & Testing & \textbf{96.22} & \textbf{94.37} & 97.25 & \textbf{99.61} & \textbf{99.75} & 99.76 \\ \hline\hline
 & Training & \textbf{100.00} & \textbf{100.00} & \textbf{100.00} & \textbf{100.00} & \textbf{100.00} & \textbf{100.00} \\ \cline{2-8} 
 & Validation & 96.10 & 93.58 & \textbf{97.68} & \textbf{99.53} & 99.64 & 99.72 \\ \cline{2-8} 
\multirow{-3}{*}{ResNet-50} & Testing & 95.67 & 93.25 & \textbf{97.42} & 99.31 & 99.61 & 99.80 \\ \hline\hline
& Training & \textbf{100.00} & \textbf{100.00} & \textbf{100.00} & \textbf{100.00} & \textbf{100.00} & \textbf{100.00} \\ \cline{2-8} 
 & Validation & 92.45 & 87.06 & 96.07 & 98.15 & 99.59 & 99.44 \\ \cline{2-8} 
 
\multirow{-3}{*}{Inception-V3} & Testing & 92.06 & 87.45 & 96.23 & 99.16 & 99.67 & \textbf{99.88} \\ \hline\hline
 & Training & \textbf{100.00} & \textbf{100.00} & \textbf{100.00} & \textbf{100.00} & \textbf{100.00} & \textbf{100.00} \\ \cline{2-8} 
 & Validation & 94.26 & 89.70 & 96.43 & 98.76 & 99.64 & 99.56 \\ \cline{2-8} 
\multirow{-3}{*}{\begin{tabular}[c]{@{}l@{}}Inception-\\ ResNet-V2\end{tabular}} & Testing & 94.25 & 90.35 & 96.93 & 99.46 & 99.67 & 99.60 \\ \hline\hline
 & Training & \textbf{100.00} & \textbf{100.00} & \textbf{100.00} & \textbf{100.00} & \textbf{100.00} & \textbf{100.00} \\ \cline{2-8} 
& Validation & 93.01 & 43.16 & 96.02 & 98.31 & 99.42 & 99.52 \\ \cline{2-8} 
\multirow{-3}{*}{MobileNetV2} 
 & Testing & 92.94 & 42.49 & 95.98 & 98.55 & 99.61 & 99.68 \\ \hline
\end{tabular}
\end{table}

\begin{figure}[t!]
    \centering
    \captionsetup[subfigure]{justification=centering}
    \begin{subfigure}[b]{.48\textwidth}
        \centering
        \includegraphics[width=\textwidth, height = 22em]{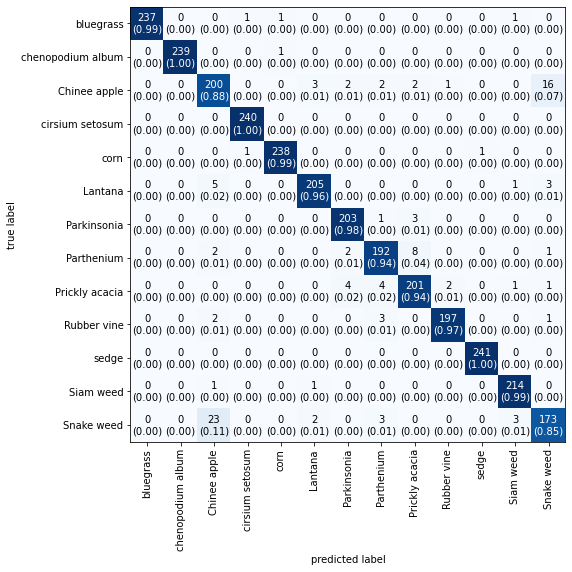}
        \caption{\scriptsize{Confusion matrix for DW-CW dataset (using VGG16 model)}}
        \label{fig:vgg_DWCW}
    \end{subfigure}
    \hfill
    \begin{subfigure}[b]{.48\textwidth}
        \centering
        \includegraphics[width=\textwidth, height = 22em]{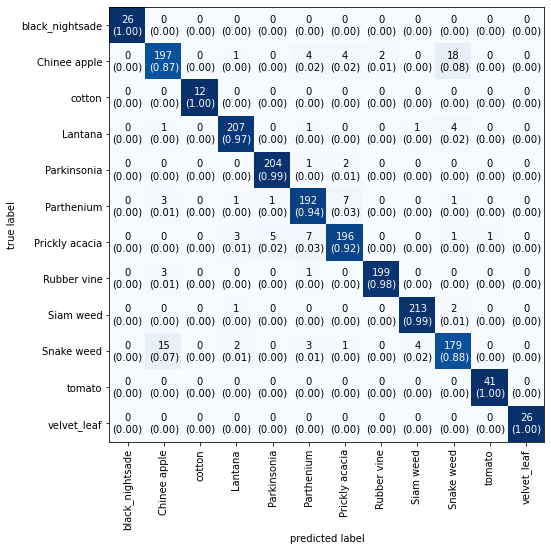}  
        \caption{\scriptsize{Confusion matrix for DW-CTW dataset (using VGG16 model)}}
        \label{fig:vgg_DWCTW}
    \end{subfigure}
    \hfill
    \hfill
    \begin{subfigure}[b]{.48\textwidth}
        \centering
        \includegraphics[width=\textwidth, height = 22em]{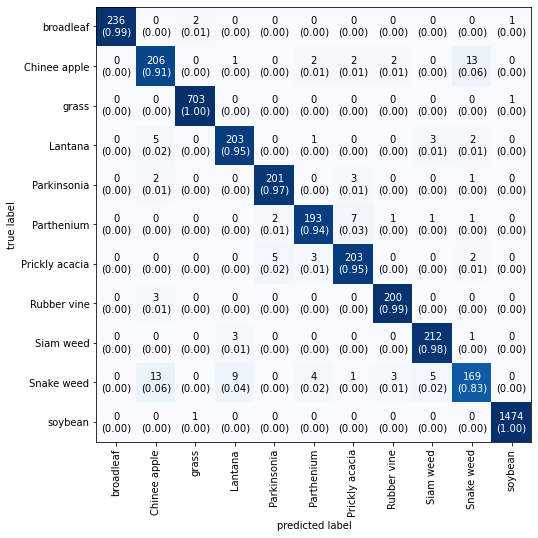}
        \caption{\scriptsize{Confusion matrix for DW-SW dataset (ResNet-50 model)}}
        \label{fig:Res_DWSW}
    \end{subfigure}
    \hfill
\caption[Confusion matrix of \enquote{DeepWeeds} combined with other three dataset.]{Confusion matrix of \enquote{DeepWeeds} combined with other three dataset.}
\label{fig:confusion_mat_DW_vs_others}
\end{figure}

The results of the confusion matrix are provided in Figure \ref{fig:confusion_mat_DW_vs_others}. We found that chinee apple, lantana, prickly acacia and snakeweed had a high confusion rate. This result agrees with that of \textcite{olsen2019deepweeds}. Visually, the images were quite similar and so were difficult to distinguish. That is why the DL model also failed to detect those. Since the dataset was small and did not have enough variations among the images, the models were not able to distinguish among the classes. The datasets also lacked enough images taken under different lighting conditions. The models were unable to detect the actual class of the images because of the illumination effects. 

\begin{figure}[t!]
    \centering
    \captionsetup[subfigure]{justification=centering}
    \begin{subfigure}[b]{.23\textwidth}
        \centering
        \includegraphics[width=\textwidth, height = 8em]{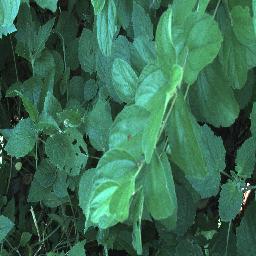}
        \caption{\scriptsize{Chinee apple predicted as snakeweed}}
        \label{fig:mis_ca}
    \end{subfigure}
    \hfill
    \begin{subfigure}[b]{.23\textwidth}
        \centering
        \includegraphics[width=\textwidth, height = 8em]{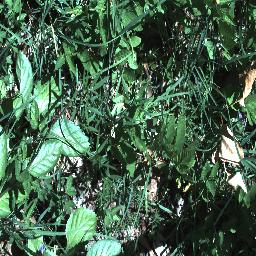}  
        \caption{\scriptsize{snakeweed predicted as chinee apple}}
        \label{fig:mis_sw}
    \end{subfigure}
    \hfill
    \begin{subfigure}[b]{.23\textwidth}
        \centering
        \includegraphics[width=\textwidth, height = 8em]{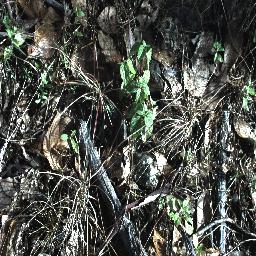}  
        \caption{\scriptsize{Lantana predicted as prickly acacia}}
        \label{fig:mis_lan}
    \end{subfigure}
    \hfill
    \begin{subfigure}[b]{.23\textwidth}
        \centering
        \includegraphics[width=\textwidth, height = 8em]{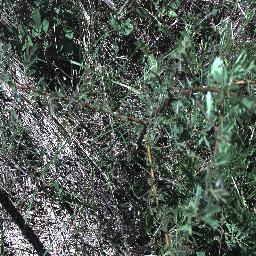}  
        \caption{\scriptsize{Prickly acacia predicted as lantana}}
        \label{fig:mis_pa}
    \end{subfigure}
    \hfill
\caption[Example of incorrectly classified images.]{Example of incorrectly classified images.}
\label{fig:false_images}
\end{figure}

For the DW-CW dataset, the VGG16 model was more accurate. In this case, the model did not distinguish between chinee apple and snakeweed. As shown in the confusion matrix in Figure \ref{fig:vgg_DWCW}, out of 224 test images of chinee apple, 16 were classified as snakeweed, and 23 of the 204 test images of snakeweed identified as chinee apple. A significant number of chinee apple and snakeweed images were not correctly predicted by the VGG16 model (see Figure \ref{fig:vgg_DWCTW}). For the DW-SW dataset, the ResNet-50 model achieved 100\% training, 97.68\% validation and  97.42\% testing accuracy. The confusion matrix is shown in Figure \ref{fig:Res_DWSW}. The ResNet-50 model identified 13 chinee apple images as snakeweed, and the same number of snakeweed images were classified as chinee apple. The model also identified 9 test images of snakeweed as lantana. Figure \ref{fig:false_images} shows some sample images which the models classified incorrectly. \par 

By applying data augmentation techniques, one can create more variations among the classes which may also help the model to learn more discriminating features. \par 

\subsection{Experiment 3: Training the model with all four datasets together} \label{tr_real}

In this experiment, all the datasets were combined to train the deep learning models. Classifying the images of the combined dataset is much more complex, as the data is highly class-imbalanced. The models were initialised with pre-trained weights and then fine-tuned. Table \ref{tab:performance_comb} shows the training, validation and testing accuracy and average precision, recall, and F1 scores achieved by the models on the test data. \par

\begin{table}[t]
\centering
\caption{The performance of five deep learning models after training with the combined dataset}
\label{tab:performance_comb}
\begin{tabular}{|l|r|r|r|r|r|r|}
\hline
\multicolumn{1}{|c|}{\textbf{DL model}}                        & \multicolumn{1}{c|}{\textbf{\begin{tabular}[c]{@{}c@{}}Training \\ accuracy\end{tabular}}} & \multicolumn{1}{c|}{\textbf{\begin{tabular}[c]{@{}c@{}}Validation \\ accuracy\end{tabular}}} & \multicolumn{1}{c|}{\textbf{\begin{tabular}[c]{@{}c@{}}Testing\\ accuracy\end{tabular}}} & \multicolumn{1}{c|}{\textbf{\begin{tabular}[c]{@{}c@{}}Precision\\ (Average)\end{tabular}}} & \multicolumn{1}{c|}{\textbf{\begin{tabular}[c]{@{}c@{}}Recall\\ (Average)\end{tabular}}} & \multicolumn{1}{c|}{\textbf{\begin{tabular}[c]{@{}c@{}}F1 score\\ (Average)\end{tabular}}} \\ \hline
VGG16 & 99.96 & 97.53 & 97.76 & 96.89 & 96.83 & 96.84 \\ \hline
ResNet-50 & \textbf{100.00} & \textbf{97.83} & \textbf{98.06} & \textbf{98.06} & \textbf{98.06} & \textbf{98.05} \\ \hline
Inception-V3 & \textbf{100.00} & 96.66 & 96.09 & 96.11 & 97.09 & 97.09 \\ \hline
\begin{tabular}[c]{@{}l@{}}Inception-\\ Resnet-V2\end{tabular} & \textbf{100.00} & 96.88 & 97.17 & 97.17 & 97.17 & 97.16 \\ \hline
MobileNetV2 & \textbf{100.00} & 96.94 & 97.17 & 97.18 & 97.17 & 97.17 \\ \hline
\end{tabular}
\end{table}

\begin{figure}[tb!]
    \centering
    \includegraphics[width=\textwidth]{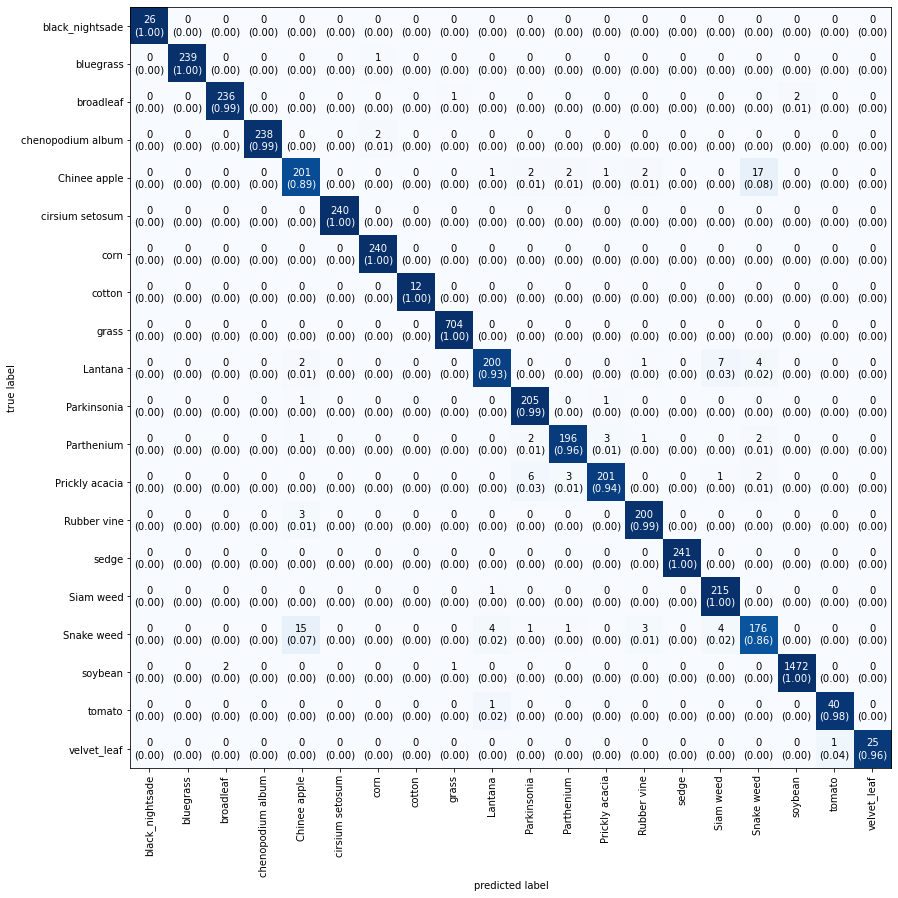}
    \caption{Confusion matrix after combining four dataset using ResNet-50 model}
    \label{fig:ResNet_four}
\end{figure}

After training the models with the combined dataset, the ResNet-50 model performed better. Though all the models except VGG16 achieved 100\% training accuracy, the validation (97.83\%) and testing (98.06\%) accuracies of ResNet-50 architecture were higher. The average precision, recall and F1 score also verified these results. However, the models still did not correctly classify the chinee apple and snakeweed species mentioned in the previous experiment (Section \ref{comb_two}).  A confusion matrix for predicting the classes of images using ResNet-50 is shown in Figure \ref{fig:ResNet_four}. The confusion of ResNet-50 is chosen, since the highest accuracy is achieved in this experiment using this model. Seventeen chinee apple images were classified as snakeweed, and fifteen snakeweeds images were classified incorrectly as chinee apple. In addition, the model also incorrectly classified some lantana and prickly acacia weed images. To overcome this classification problem, both actual and augmented data were used in the following experiment. \par

\subsection{Experiment 4: Training the models using both real and augmented images of the four datasets}

Augmented data were used together with the real data in the training phase to address the misclassification problem in the previous experiment (Section \ref{tr_real}). All the weed species and crop plant images had the same training data for this experiment. The models were initialised with pre-trained weights, and all the parameters were fine-tuned. Table \ref{tab:per_real_aug} shows the result of this experiment. \par

\begin{table}[]
\centering
\caption{Performance of five deep learning models after training with the real and augmented data}
\label{tab:per_real_aug}
\begin{tabular}{|l|r|r|r|r|r|r|}
\hline
\multicolumn{1}{|c|}{\textbf{DL model}} & \multicolumn{1}{c|}{\textbf{\begin{tabular}[c]{@{}c@{}}Training \\ accuracy\end{tabular}}} & \multicolumn{1}{c|}{\textbf{\begin{tabular}[c]{@{}c@{}}Validation\\ accuracy\end{tabular}}} & \multicolumn{1}{c|}{\textbf{\begin{tabular}[c]{@{}c@{}}Testing\\ accuracy\end{tabular}}} & \multicolumn{1}{c|}{\textbf{\begin{tabular}[c]{@{}c@{}}Precision\\ (Average)\end{tabular}}} & \multicolumn{1}{c|}{\textbf{\begin{tabular}[c]{@{}c@{}}Recall\\ (Average)\end{tabular}}} & \multicolumn{1}{c|}{\textbf{\begin{tabular}[c]{@{}c@{}}F1 score\\ (Average)\end{tabular}}} \\ \hline
VGG16 & 100.00 & 97.96 & 97.83 & 97.83 & 97.84 & 97.83 \\ \hline
ResNet-50 & \textbf{100.00} & \textbf{98.31} & \textbf{98.30} & \textbf{98.29} & \textbf{98.30} & \textbf{98.30} \\ \hline
Inception-V3 & 100.00 & 97.31 & 98.02 & 98.02 & 98.02 & 98.01 \\ \hline
Inception-Resnet-V2 & 100.00 & 97.85 & 97.76 & 97.76 & 97.76 & 97.76 \\ \hline
MobileNetV2 & 100.00 & 97.68 & 98.02 & 98.02 & 98.02 & 98.02 \\ \hline
\end{tabular}
\end{table}

\begin{figure}[tb!]
    \centering
    \includegraphics[width=\textwidth]{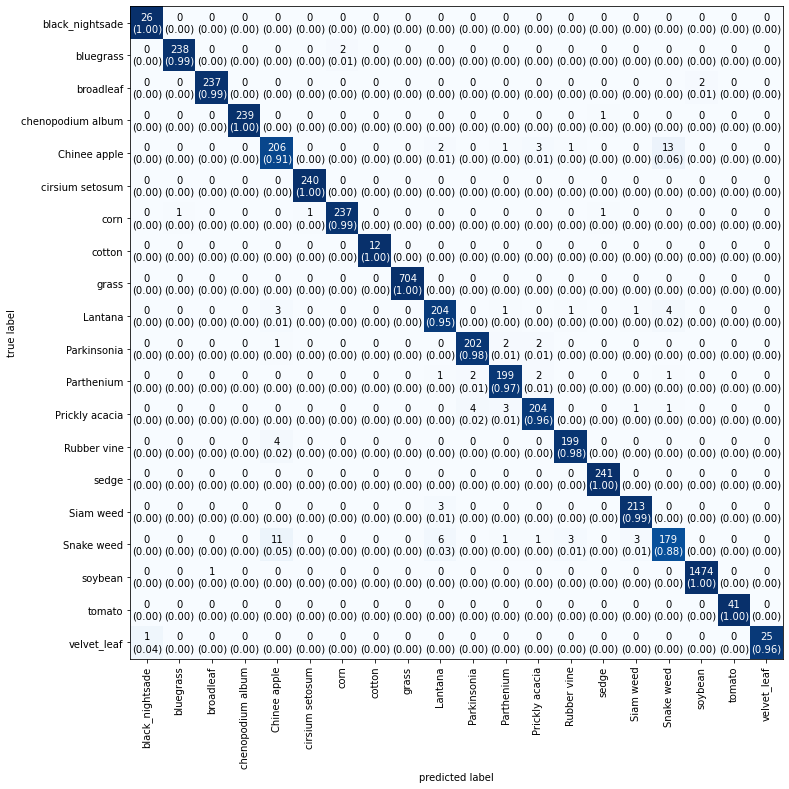}
    \caption{Confusion matrix for ResNet-50 model using combined dataset with augmentation}
    \label{fig:ResNet_four_aug}
\end{figure}

From Table \ref{tab:per_real_aug}, we can see that the training accuracy for all the DL models is 100\%. Also the validation and testing accuracies were reasonably high. In this experiment, the ResNet-50 models achieved the highest precision, recall and F1 score for the test data. Figure \ref{fig:ResNet_four_aug} shows the confusion matrix for the ResNet-50 model. We compared the performance of the model using the confusion matrix with the previous experiment. The performance of the model was improved using both actual and augmented data. The classification accuracy increased for chinee apple, lantana, prickly acacia and snakeweed species by 2\%. \par

In this research, the ResNet-50 model attained the highest accuracy using actual and augmented images. The Inception-ResNet-V2 model gave similar results. The explanation is that both of the models used residual layers. Residual connections help train a deeper neural network with better performance and reduced computational complexity. A deeper convolutional network works better when trained using a large dataset \parencite{szegedy2017inception}. Since we have used the augmented data and actual images, the dataset size has increased by several times. \par

\subsection{Experiment 5: Comparing the performance of two ResNet-50 models individually trained on ImageNet dataset, and the combined dataset, and testing on the Unseen Test dataset}

\begin{figure}[ht!]
    \centering
    \captionsetup[subfigure]{justification=centering}
    
    \begin{subfigure}[b]{.45\textwidth}
        \centering
        \includegraphics[width=\textwidth, height = 20em]{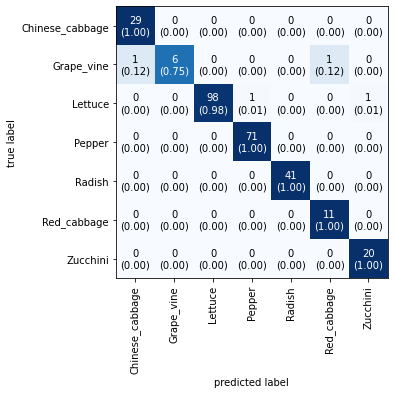}  
        \caption{\scriptsize{Confusion matrix showing the classification accuracy of CW ResNet-50 model}}
        \label{fig:ResNet_CW}
    \end{subfigure}
    \hfil
    \begin{subfigure}[b]{.45\textwidth}
        \centering
        \includegraphics[width=\textwidth, height = 20em]{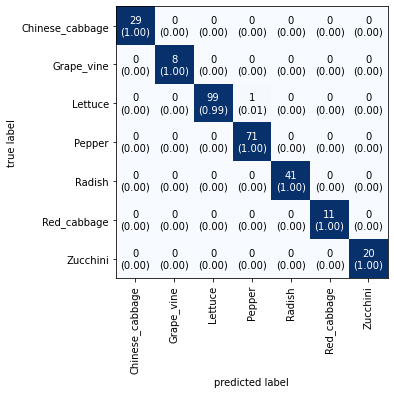}  
        \caption{\scriptsize{Confusion matrix showing the classification accuracy of CW ResNet-50 model}}
        \label{fig:ResNet_SOTA}
    \end{subfigure}
    \hfil

\caption[Confusion matrix for CW ResNet-50 and SOTA ResNet-50 model.]{Confusion matrix for CW ResNet-50 and SOTA ResNet-50 model.}
\label{fig:CW_SOTA_Res}
\end{figure}

In this experiment, we used two ResNet-50 models. The first was trained on our combined dataset with actual and augmented data (Sec. \ref{ComDataset}).  Here, the top layers were removed from the model and a global average pooling layer and three dense layers were added as before. Other than the top layers, all the layers used pre-trained weights, which were not fine-tuned. This model termed as \enquote{CW ResNet-50}. The same arrangement was used for the pre-trained ResNet-50 model, which was instead trained on the ImageNet dataset. It was named as \enquote{SOTA ResNet-50} model for further use. We trained the top layers of both models using the training split of the Unseen Test Dataset (\ref{UnseenTestData}). Both models were tested using the test split of the Unseen Test Dataset. The confusion matrix for CW ResNet-50 and SOTA ResNet-50 model is shown in Figure \ref{fig:CW_SOTA_Res}. \par

We can see in Figure \ref{fig:CW_SOTA_Res} that the performance of the two models is very similar. The \enquote{SOTA ResNet-50} model detected all the classes of crop and weeds accurately. However, the pre-trained \enquote{CW Resnet-50} model only identified two images incorrectly. As the \enquote{SOTA ResNet-50} model was trained on a large dataset containing millions of images, it detected the discriminating features more accurately. On the other hand, the \enquote{CW Resnet-50} model was only trained on 88,500 images. If this model were trained with more data, it is probable that it would be more accurate using the transfer learning approach. This type of pre-trained model could be used for classifying the images of new crop and weed datasets, which would eventually make the training process faster. \par

\section{Conclusion} \label{conclusion}
This study was undertaken on four image datasets of crop and weed species collected from four different geographical locations. The datasets contained a total of 20 different species of crops and weeds. We used five state-of-the-art CNN models, namely VGG16, ResNet-50, Inception-V3, Inception- ResNet-V2, MobileNetV2, to classify the images of these crops and weeds.\par

First, we evaluated the performance of transfer learning and fine-tuning approaches of the models by training them on each dataset. The results showed that fine-tuning of the models could improve classification of the images more accurately than the transfer learning approach. \par

To add more complexity to the classification problem, we combined the datasets together. After combining two of the datasets, the performance decreased due to some of the species of weeds in the \enquote{DeepWeeds} dataset. The weed species that were confused were chinee apple, snakeweed, lantana and prickly acacia. We then combined all four datasets to train the models. Since the dataset was class-imbalanced, it was difficult to achieve high classification accuracy by only training the model with actual images. Consequently, we used augmentation to balance the classes of the dataset. However, it was evident that the models had problems in distinguishing between chinee apple and snakeweed. The performance of the models improved using both actual and augmented data. The models could distinguish chinee apple and snake weed more accurately. The results showed that the ResNet-50 was most accurate.\par

Another finding was that using the transfer learning method was that in most cases the models did not achieve the desired accuracy. As ResNet-50 was the most accurate system, we ran a test using this pre-trained model. The model was used to classify the images of a new dataset using the transfer learning approach. Although the model was not more accurate than the state-of-the-art pre-trained ResNet-50 model, it was very close to that. We could expect a higher accuracy using the transfer learning approach if the model can be trained using a large crop and weed dataset.\par

This research shows that the data augmentation technique can help address the class imbalance problem and add more variations to the dataset. The variations in the images of the training dataset improve the training accuracy of the deep learning models. Moreover, the transfer learning approach can mitigate the requirement of large data sets to train the deep learning models from scratch. The pre-trained models are trained on a large dataset to capture the detailed generalised features from the imagery, e.g., ImageNet in our case. However, because, ImageNet data set was not categorically labelled for weeds or crops, fine-tuning the pre-trained weights with crop and weed datasets help capture the dataset or task-specific features. Consequently, fine-tuning improves classification accuracy. 

For training a deep learning model for classifying images, it is essential to have a large dataset like ImageNet \parencite{deng2009imagenet} and MS-COCO \parencite{lin2014microsoft}. Classification of crop and weed species cannot be generalised unless a benchmark dataset is available. Most studies in this area are site-specific. A large dataset is needed to generalise the classification of crop and weed plants, and as an initial approach, large datasets can be generated by combining multiple small datasets, as demonstrated here. In this work, the images only had image-level labels. A benchmark dataset can be created by combining many datasets annotated with a variety of image labelling techniques. Generative Adversarial Networks (GANs) \parencite{goodfellow2014generative} based image sample generation can also be used to mitigate class-imbalance issues. Moreover, it is needed to develop a crop \& weed dataset annotated at the object level. For implementing a real-time selective herbicide sprayer, the classification of weed species is not enough. It is also necessary to locate the weeds in crops. Deep learning-based object detection models can be used for detecting weeds. \par

\section{Declaration of Funding}
This research did not receive any specific funding.

\section{Conflicts of interest}
The authors declare no conflict of interest.

\section{Data availability}
The data that support this study will be shared upon reasonable request to the corresponding author.

\cleardoublepage

\printbibliography

@article{dargan2019survey,
  title={A survey of deep learning and its applications: A new paradigm to machine learning},
  author={Dargan, Shaveta and Kumar, Munish and Ayyagari, Maruthi Rohit and Kumar, Gulshan},
  journal={Archives of Computational Methods in Engineering},
  pages={1--22},
  year={2019},
  publisher={Springer}
}

@misc{steinberg_2017, 
title={6 areas where artificial neural networks outperform humans},
url={https://venturebeat.com/2017/12/08/6-areas-where-artificial-neural-networks-outperform-humans/},
journal={VentureBeat}, 
publisher={VentureBeat}, 
author={Steinberg, Roman}, 
year={2017}, 
month={12},
urldate = {2020-12-25}}

@article{kamilaris2018deep,
  title={Deep learning in agriculture: A survey},
  author={Kamilaris, Andreas and Prenafeta-Bold{\'u}, Francesc X},
  journal={Computers and electronics in agriculture},
  volume={147},
  pages={70--90},
  year={2018},
  publisher={Elsevier}
}

@article{iqbal2019investigation,
  title={Investigation of alternate herbicides for effective weed management in glyphosate-tolerant cotton},
  author={Iqbal, Nadeem and Manalil, Sudheesh and Chauhan, Bhagirath S and Adkins, Steve W},
  journal={Archives of Agronomy and Soil Science},
  volume={65},
  number={13},
  pages={1885--1899},
  year={2019},
  publisher={Taylor \& Francis}
}

@inproceedings{lameski2018review,
  title={Review of automated weed control approaches: an environmental impact perspective},
  author={Lameski, Petre and Zdravevski, Eftim and Kulakov, Andrea},
  booktitle={International Conference on Telecommunications},
  pages={132--147},
  year={2018},
  organization={Springer}
}

@article{slaughter2008autonomous,
  title={Autonomous robotic weed control systems: A review},
  author={Slaughter, DC and Giles, DK and Downey, D},
  journal={Computers and electronics in agriculture},
  volume={61},
  number={1},
  pages={63--78},
  year={2008},
  publisher={Elsevier}
}

@article{olsen2019deepweeds,
  title={DeepWeeds: A multiclass weed species image dataset for deep learning},
  author={Olsen, Alex and Konovalov, Dmitry A and Philippa, Bronson and Ridd, Peter and Wood, Jake C and Johns, Jamie and Banks, Wesley and Girgenti, Benjamin and Kenny, Owen and Whinney, James and others},
  journal={Scientific reports},
  volume={9},
  number={1},
  pages={1--12},
  year={2019},
  publisher={Nature Publishing Group}
}

@article{waldchen2018plant,
  title={Plant species identification using computer vision techniques: A systematic literature review},
  author={W{\"a}ldchen, Jana and M{\"a}der, Patrick},
  journal={Archives of Computational Methods in Engineering},
  volume={25},
  number={2},
  pages={507--543},
  year={2018},
  publisher={Springer}
}

@article{chavan2018agroavnet,
  title={AgroAVNET for crops and weeds classification: A step forward in automatic farming},
  author={Chavan, Trupti R and Nandedkar, Abhijeet V},
  journal={Computers and Electronics in Agriculture},
  volume={154},
  pages={361--372},
  year={2018},
  publisher={Elsevier}
}

@article{teimouri2018weed,
  title={Weed growth stage estimator using deep convolutional neural networks},
  author={Teimouri, Nima and Dyrmann, Mads and Nielsen, Per Rydahl and Mathiassen, Solvejg Kopp and Somerville, Gayle J and J{\o}rgensen, Rasmus Nyholm},
  journal={Sensors},
  volume={18},
  number={5},
  pages={1580},
  year={2018},
  publisher={Multidisciplinary Digital Publishing Institute}
}

@article{peteinatos2020weed,
  title={Weed Identification in Maize, Sunflower, and Potatoes with the Aid of Convolutional Neural Networks},
  author={Peteinatos, Gerassimos and Reichel, Philipp and Karouta, Jeremy and And{\'u}jar, Dionisio and Gerhards, Roland},
  journal={Remote Sensing},
  volume={12},
  number={24},
  pages={4185},
  year={2020},
  publisher={Multidisciplinary Digital Publishing Institute}
}

@article{dyrmann2016plant,
  title={Plant species classification using deep convolutional neural network},
  author={Dyrmann, Mads and Karstoft, Henrik and Midtiby, Henrik Skov},
  journal={Biosystems Engineering},
  volume={151},
  pages={72--80},
  year={2016},
  publisher={Elsevier}
}

@article{ali2019mfc,
  title={MFC-GAN: class-imbalanced dataset classification using multiple fake class generative adversarial network},
  author={Ali-Gombe, Adamu and Elyan, Eyad},
  journal={Neurocomputing},
  volume={361},
  pages={212--221},
  year={2019},
  publisher={Elsevier}
}

@article{krawczyk2016learning,
  title={Learning from imbalanced data: open challenges and future directions},
  author={Krawczyk, Bartosz},
  journal={Progress in Artificial Intelligence},
  volume={5},
  number={4},
  pages={221--232},
  year={2016},
  publisher={Springer}
}

@article{suh2018transfer,
  title={Transfer learning for the classification of sugar beet and volunteer potato under field conditions},
  author={Suh, Hyun K and Ijsselmuiden, Joris and Hofstee, Jan Willem and van Henten, Eldert J},
  journal={Biosystems engineering},
  volume={174},
  pages={50--65},
  year={2018},
  publisher={Elsevier}
}

@article{dos2017weed,
  title={Weed detection in soybean crops using ConvNets},
  author={Ferreira, Alessandro dos Santos and Freitas, Daniel Matte and da Silva, Gercina Gon{\c{c}}alves and Pistori, Hemerson and Folhes, Marcelo Theophilo},
  journal={Computers and Electronics in Agriculture},
  volume={143},
  pages={314--324},
  year={2017},
  publisher={Elsevier}
}

@article{nkemelu2018deep,
  title={Deep convolutional neural network for plant seedlings classification},
  author={Nkemelu, Daniel K and Omeiza, Daniel and Lubalo, Nancy},
  journal={arXiv preprint arXiv:1811.08404},
  year={2018}
}

@article{espejo2020towards,
  title={Towards weeds identification assistance through transfer learning},
  author={Espejo-Garcia, Borja and Mylonas, Nikos and Athanasakos, Loukas and Fountas, Spyros and Vasilakoglou, Ioannis},
  journal={Computers and Electronics in Agriculture},
  volume={171},
  pages={105306},
  year={2020},
  publisher={Elsevier}
}

@article{jiang2020cnn,
  title={CNN feature based graph convolutional network for weed and crop recognition in smart farming},
  author={Jiang, Honghua and Zhang, Chuanyin and Qiao, Yongliang and Zhang, Zhao and Zhang, Wenjing and Song, Changqing},
  journal={Computers and Electronics in Agriculture},
  volume={174},
  pages={105450},
  year={2020},
  publisher={Elsevier}
}

@online{chollet2015keras,
  title={Keras},
  author={Chollet, Francois and others},
  year={2015},
  publisher={GitHub},
  url={https://github.com/fchollet/keras},
}

@inproceedings{abadi2016tensorflow,
  title={Tensorflow: A system for large-scale machine learning},
  author={Abadi, Mart{\'\i}n and Barham, Paul and Chen, Jianmin and Chen, Zhifeng and Davis, Andy and Dean, Jeffrey and Devin, Matthieu and Ghemawat, Sanjay and Irving, Geoffrey and Isard, Michael and others},
  booktitle={12th $\{$USENIX$\}$ Symposium on Operating Systems Design and Implementation ($\{$OSDI$\}$ 16)},
  pages={265--283},
  year={2016}
}

@article{simonyan2014very,
  title={Very deep convolutional networks for large-scale image recognition},
  author={Simonyan, Karen and Zisserman, Andrew},
  journal={arXiv preprint arXiv:1409.1556},
  year={2014}
}

@inproceedings{he2016deep,
  title={Deep residual learning for image recognition},
  author={He, Kaiming and Zhang, Xiangyu and Ren, Shaoqing and Sun, Jian},
  booktitle={Proceedings of the IEEE conference on computer vision and pattern recognition},
  pages={770--778},
  year={2016}
}

@inproceedings{szegedy2016rethinking,
  title={Rethinking the inception architecture for computer vision},
  author={Szegedy, Christian and Vanhoucke, Vincent and Ioffe, Sergey and Shlens, Jon and Wojna, Zbigniew},
  booktitle={Proceedings of the IEEE conference on computer vision and pattern recognition},
  pages={2818--2826},
  year={2016}
}

@inproceedings{szegedy2017inception,
  title={Inception-v4, inception-resnet and the impact of residual connections on learning},
  author={Szegedy, Christian and Ioffe, Sergey and Vanhoucke, Vincent and Alemi, Alexander},
  booktitle={Proceedings of the AAAI Conference on Artificial Intelligence},
  volume={31},
  number={1},
  year={2017}
}

@inproceedings{sandler2018mobilenetv2,
  title={Mobilenetv2: Inverted residuals and linear bottlenecks},
  author={Sandler, Mark and Howard, Andrew and Zhu, Menglong and Zhmoginov, Andrey and Chen, Liang-Chieh},
  booktitle={Proceedings of the IEEE conference on computer vision and pattern recognition},
  pages={4510--4520},
  year={2018}
}

@article{van2014scikit,
  title={scikit-image: image processing in Python},
  author={Van der Walt, Stefan and Sch{\"o}nberger, Johannes L and Nunez-Iglesias, Juan and Boulogne, Fran{\c{c}}ois and Warner, Joshua D and Yager, Neil and Gouillart, Emmanuelle and Yu, Tony},
  journal={PeerJ},
  volume={2},
  pages={e453},
  year={2014},
  publisher={PeerJ Inc.}
}

@inproceedings{deng2009imagenet,
  title={Imagenet: A large-scale hierarchical image database},
  author={Deng, Jia and Dong, Wei and Socher, Richard and Li, Li-Jia and Li, Kai and Fei-Fei, Li},
  booktitle={2009 IEEE conference on computer vision and pattern recognition},
  pages={248--255},
  year={2009},
  organization={Ieee}
}

@inproceedings{lin2014microsoft,
  title={Microsoft coco: Common objects in context},
  author={Lin, Tsung-Yi and Maire, Michael and Belongie, Serge and Hays, James and Perona, Pietro and Ramanan, Deva and Doll{\'a}r, Piotr and Zitnick, C Lawrence},
  booktitle={European conference on computer vision},
  pages={740--755},
  year={2014},
  organization={Springer}
}

@article{khan2017cost,
  title={Cost-sensitive learning of deep feature representations from imbalanced data},
  author={Khan, Salman H and Hayat, Munawar and Bennamoun, Mohammed and Sohel, Ferdous A and Togneri, Roberto},
  journal={IEEE transactions on neural networks and learning systems},
  volume={29},
  number={8},
  pages={3573--3587},
  year={2017},
  publisher={IEEE}
}

@article{achanta2012slic,
  title={SLIC superpixels compared to state-of-the-art superpixel methods},
  author={Achanta, Radhakrishna and Shaji, Appu and Smith, Kevin and Lucchi, Aurelien and Fua, Pascal and S{\"u}sstrunk, Sabine},
  journal={IEEE transactions on pattern analysis and machine intelligence},
  volume={34},
  number={11},
  pages={2274--2282},
  year={2012},
  publisher={IEEE}
}

@article{sabottke2020effect,
  title={The effect of image resolution on deep learning in radiography},
  author={Sabottke, Carl F and Spieler, Bradley M},
  journal={Radiology: Artificial Intelligence},
  volume={2},
  number={1},
  pages={e190015},
  year={2020},
  publisher={Radiological Society of North America}
}

@article{sahlsten2019deep,
  title={Deep learning fundus image analysis for diabetic retinopathy and macular edema grading},
  author={Sahlsten, Jaakko and Jaskari, Joel and Kivinen, Jyri and Turunen, Lauri and Jaanio, Esa and Hietala, Kustaa and Kaski, Kimmo},
  journal={Scientific reports},
  volume={9},
  number={1},
  pages={1--11},
  year={2019},
  publisher={Nature Publishing Group}
}

@inproceedings{guo2019spottune,
  title={Spottune: transfer learning through adaptive fine-tuning},
  author={Guo, Yunhui and Shi, Honghui and Kumar, Abhishek and Grauman, Kristen and Rosing, Tajana and Feris, Rogerio},
  booktitle={Proceedings of the IEEE/CVF Conference on Computer Vision and Pattern Recognition},
  pages={4805--4814},
  year={2019}
}

@article{pan2009survey,
  title={A survey on transfer learning},
  author={Pan, Sinno Jialin and Yang, Qiang},
  journal={IEEE Transactions on knowledge and data engineering},
  volume={22},
  number={10},
  pages={1345--1359},
  year={2009},
  publisher={IEEE}
}

@article{goodfellow2014generative,
  title={Generative adversarial networks},
  author={Goodfellow, Ian J and Pouget-Abadie, Jean and Mirza, Mehdi and Xu, Bing and Warde-Farley, David and Ozair, Sherjil and Courville, Aaron and Bengio, Yoshua},
  journal={arXiv preprint arXiv:1406.2661},
  year={2014}
}

@misc{machinery_2018, title={Robocrop Spot Sprayer: Weed Removal}, url={https://garford.com/products/robocrop-spot-sprayer/}, journal={Garford Farm Machinery}, year={2018}, month={7}, urldate = {2021-01-25}}

@misc{trimble_agriculture, title={WeedSeeker 2 Spot Spray System}, url={https://agriculture.trimble.com/product/weedseeker-2-spot-spray-system/}, journal={Trimble Agriculture}, urldate = {2021-01-25}}

@misc{weed, title={Precision Spraying - Weed Sprayer}, url={https://www.weed-it.com/}, journal={WEED}, urldate = {2021-01-25}}

@article{tian2020computer,
  title={Computer vision technology in agricultural automation—A review},
  author={Tian, Hongkun and Wang, Tianhai and Liu, Yadong and Qiao, Xi and Li, Yanzhou},
  journal={Information Processing in Agriculture},
  volume={7},
  number={1},
  pages={1--19},
  year={2020},
  publisher={Elsevier}
}

@inproceedings{lameski2017weed,
  title={Weed detection dataset with RGB images taken under variable light conditions},
  author={Lameski, Petre and Zdravevski, Eftim and Trajkovik, Vladimir and Kulakov, Andrea},
  booktitle={International Conference on ICT Innovations},
  pages={112--119},
  year={2017},
  organization={Springer}
}

@article{hasan2021survey,
  title={A survey of deep learning techniques for weed detection from images},
  author={Hasan, ASM Mahmudul and Sohel, Ferdous and Diepeveen, Dean and Laga, Hamid and Jones, Michael GK},
  journal={Computers and Electronics in Agriculture},
  volume={184},
  pages={106--067},
  year={2021},
  publisher={Elsevier}
}

@article{shao2014transfer,
  title={Transfer learning for visual categorization: A survey},
  author={Shao, Ling and Zhu, Fan and Li, Xuelong},
  journal={IEEE transactions on neural networks and learning systems},
  volume={26},
  number={5},
  pages={1019--1034},
  year={2014},
  publisher={IEEE}
}

@inproceedings{girshick2014rich,
  title={Rich feature hierarchies for accurate object detection and semantic segmentation},
  author={Girshick, Ross and Donahue, Jeff and Darrell, Trevor and Malik, Jitendra},
  booktitle={Proceedings of the IEEE conference on computer vision and pattern recognition},
  pages={580--587},
  year={2014}
}

@inproceedings{hentschel2016fine,
  title={Fine tuning CNNS with scarce training data—Adapting ImageNet to art epoch classification},
  author={Hentschel, Christian and Wiradarma, Timur Pratama and Sack, Harald},
  booktitle={2016 IEEE International Conference on Image Processing (ICIP)},
  pages={3693--3697},
  year={2016},
  organization={IEEE}
}

@article{gando2016fine,
  title={Fine-tuning deep convolutional neural networks for distinguishing illustrations from photographs},
  author={Gando, Gota and Yamada, Taiga and Sato, Haruhiko and Oyama, Satoshi and Kurihara, Masahito},
  journal={Expert Systems with Applications},
  volume={66},
  pages={295--301},
  year={2016},
  publisher={Elsevier}
}

@article{ahmad2021performance,
  title={Performance of deep learning models for classifying and detecting common weeds in corn and soybean production systems},
  author={Ahmad, Aanis and Saraswat, Dharmendra and Aggarwal, Varun and Etienne, Aaron and Hancock, Benjamin},
  journal={Computers and Electronics in Agriculture},
  volume={184},
  pages={106081},
  year={2021},
  publisher={Elsevier}
}

@article{jensen2020automated,
  title={An automated site-specific fallow weed management system using unmanned aerial vehicles},
  author={Jensen, Troy Arnold and Smith, Bruen and Defeo, Livia Faria},
  year={2020}
}

@misc{mcleod_2018, title={Annual Costs of Weeds in Australia}, url={https://invasives.com.au/wp-content/uploads/2019/01/Cost-of-weeds-report.pdf}, journal={ Centre for Invasive Species Solutions}, publisher={eSYS Development Pty Limited}, author={McLeod , Ross}, year={2018}, month={11}}

@article{harker2013recent,
  title={Recent weed control, weed management, and integrated weed management},
  author={Harker, K Neil and O'Donovan, John T},
  journal={Weed Technology},
  volume={27},
  number={1},
  pages={1--11},
  year={2013},
  publisher={Cambridge University Press}
}

@article{lopez2011weed,
  title={Weed detection for site-specific weed management: mapping and real-time approaches},
  author={L{\'O}PEZ-GRANADOS, Francisca},
  journal={Weed Research},
  volume={51},
  number={1},
  pages={1--11},
  year={2011},
  publisher={Wiley Online Library}
}

@article{european20202018,
  title={The 2018 European Union report on pesticide residues in food},
  author={Medina-Pastor, Paula and Triacchini, Giuseppe},
  journal={EFSA Journal},
  volume={18},
  number={4},
  pages={e06057},
  year={2020},
  publisher={Wiley Online Library}
}

\end{document}